\documentclass{article}

\usepackage[preprint]{neurips_2025}

\usepackage{graphicx}
\usepackage{subcaption}
\usepackage{float}

\usepackage{booktabs}    
\usepackage{array}
\usepackage{tabularx} 
\usepackage{rotating}

\newcolumntype{H}{>{\centering\arraybackslash\bfseries}m{2.2cm}}

\newcolumntype{C}{>{\centering\arraybackslash}X}

\newcommand{\imgcell}[1]{\includegraphics[width=1.5cm]{#1}}

\usepackage[utf8]{inputenc} 
\usepackage[T1]{fontenc}    
\usepackage{hyperref}       
\usepackage{url}            
\usepackage{booktabs}       
\usepackage{amsfonts}       
\usepackage{nicefrac}       
\usepackage{microtype}      
\usepackage{xcolor}         
\usepackage{wrapfig}
\usepackage{graphicx}
\usepackage{amsmath}
\usepackage{hyperref}

\usepackage{enumitem}      
\setlist{nosep}

\usepackage{amssymb}    
\usepackage{multirow}           
\usepackage{algorithm}          
\usepackage{algpseudocode}
\usepackage[title]{appendix}

\title{MatPredict: a dataset and benchmark for learning material properties of diverse indoor objects}

\author{%
  Yuzhen Chen \And Hojun Son  \And Arpan Kusari \\
  \And
  University of Michigan Transportation Research Institute \\
  University of Michigan \\
  Ann Arbor, MI 48109 \\
  \texttt{\{yuzhench, hojunson, kusari\}@umich.edu} \\
}

\begin{document}

\maketitle

\begin{abstract}
Determining material properties from camera images can expand the ability to identify complex objects in indoor environments, which is valuable for consumer robotics applications. To support this, we introduce MatPredict, a dataset that combines the high-quality synthetic objects from Replica dataset with MatSynth dataset's material properties classes - to create objects with diverse material properties. We select 3D meshes of specific foreground objects and render them with different material properties. In total, we generate \textbf{18} commonly occurring objects with \textbf{14} different materials. We showcase how we provide variability in terms of lighting and camera placement for these objects. Next, we provide a benchmark for inferring material properties from visual images using these perturbed models in the scene, discussing the specific neural network models involved and their performance based on different image comparison metrics. By accurately simulating light interactions with different materials, we can enhance realism, which is crucial for training models effectively through large-scale simulations. This research aims to revolutionize perception in consumer robotics. The dataset is provided \href{https://huggingface.co/datasets/UMTRI/MatPredict}{here} and the code is provided \href{https://github.com/arpan-kusari/MatPredict}{here}.
\end{abstract}

\section{Introduction}
\begin{wrapfigure}{r}{0.35\textwidth}
    \centering
    \includegraphics[width=0.8\linewidth]{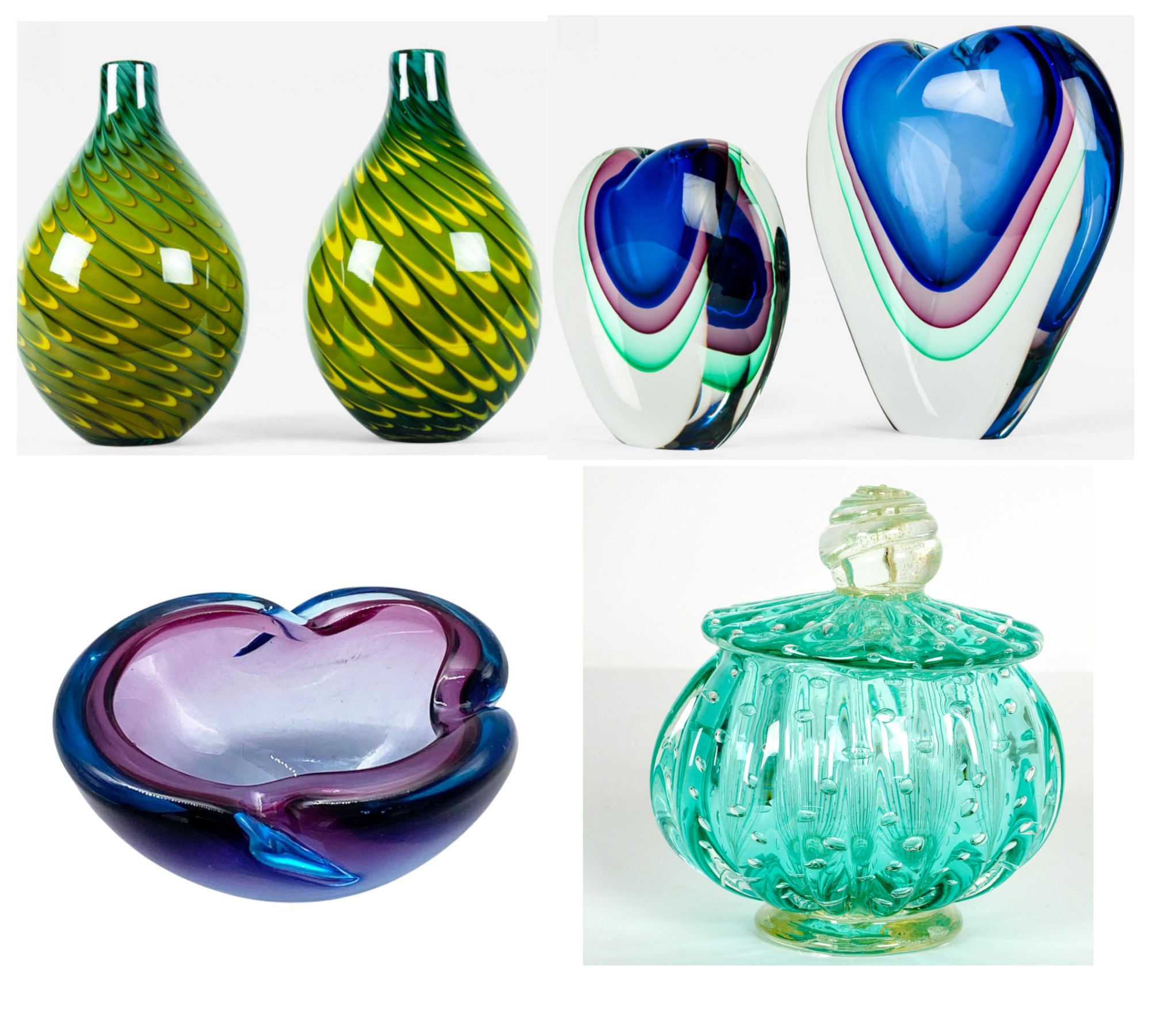} 
    \caption{Example glass decorative pieces}
    \label{fig:glass}
\end{wrapfigure}
Material properties through visual identification form a reliable way of interacting with unknown objects in the real world. For example, identification of fragile items helps determine the force and the touch points when handled by robots (different examples of glass items shown in Fig. \ref{fig:glass}). Going beyond the fragile items, material properties can refer to visual properties such as glossiness or translucency, as well as physical or tactile properties such as hardness or roughness \cite{schwartz2019recognizing}. 
Humans are remarkably good at identifying numerous different categories of materials: textiles, stones, liquids and further recognize specific materials within each class such as silk, wool and cotton \cite{fleming2014visual}. Previous research has demonstrated that subjects can make precise judgment in 
by inferring material properties such as hardness, glossiness and prettiness from photographs only \cite{fleming2013perceptual}. 
There is a growing body of experimental evidence that humans usually have an acute sense of the ``look and feel'' of an unknown object before we touch the object. 
Our aim through this proposed research is to emulate this high-level understanding of material property through training deep neural networks.

Physically based rendering (PBR) has been proposed as a way to perform image synthesis by stressing on the physical correctness of the rendering. It can be defined as:
\begin{equation}
    L_o(x,V) = L_e(x,V) + \sum_n f_r(x, L_n, V) L_i(x, L_n) (L_n \cdot N)
\end{equation}
where $L_o(x,V)$ is the outgoing light to the camera from fragment position $x$ and view vector $V$, $L_e(x,V)$ is the emitted light of the object, $n$ is the number of light sources, $f_r(x, L_n, V)$ is the bidirectional reflectance distribution function (BRDF) which provides the material properties (such as base color of the surface, metallicness, roughness, fresnel reflectance, anisotropy and transmission), $L_i(x, L_n)$ represents the incoming light and $L_n \cdot N$ represents the dot product between the light and the surface normal vector. We utilize the PBR equation to learn the material properties of the objects in a scene (also known as inverse rendering).

\begin{figure}[H]
  \centering
  \begin{tabular}{ccc}
    \begin{subfigure}[b]{0.2\textwidth}
      \centering
      \includegraphics[width=\textwidth]{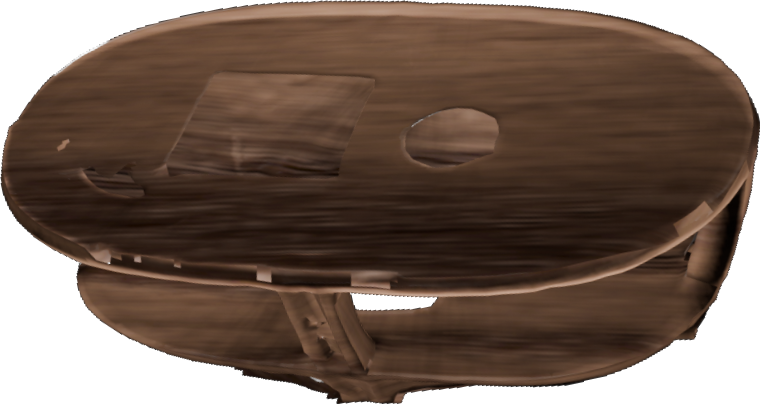}
      \caption{Wood}
    \end{subfigure}
    &
    \begin{subfigure}[b]{0.2\textwidth}
      \centering
      \includegraphics[width=\textwidth]{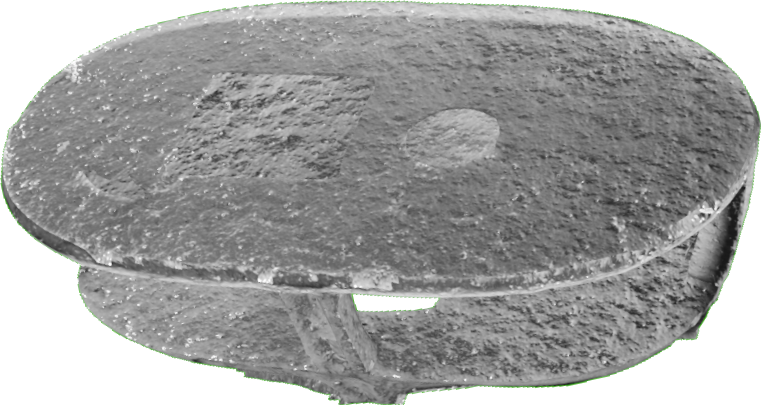}
      \caption{Concrete}
    \end{subfigure}
    &
    \begin{subfigure}[b]{0.2\textwidth}
      \centering
      \includegraphics[width=\textwidth]{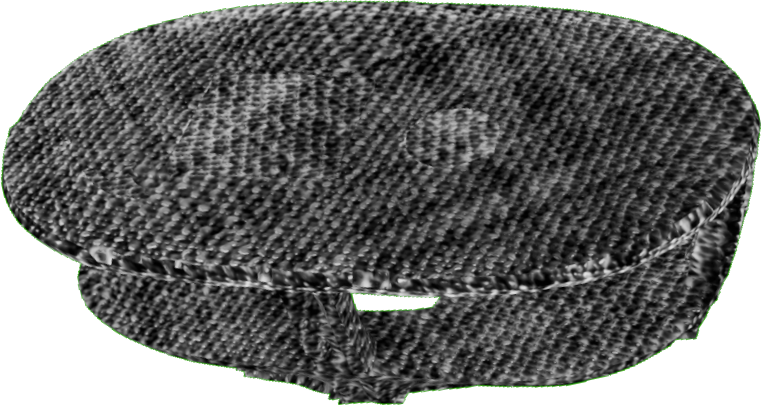}
      \caption{Fabric}
    \end{subfigure}
    \\[1em]
    \begin{subfigure}[b]{0.2\textwidth}
      \centering
      \includegraphics[width=\textwidth]{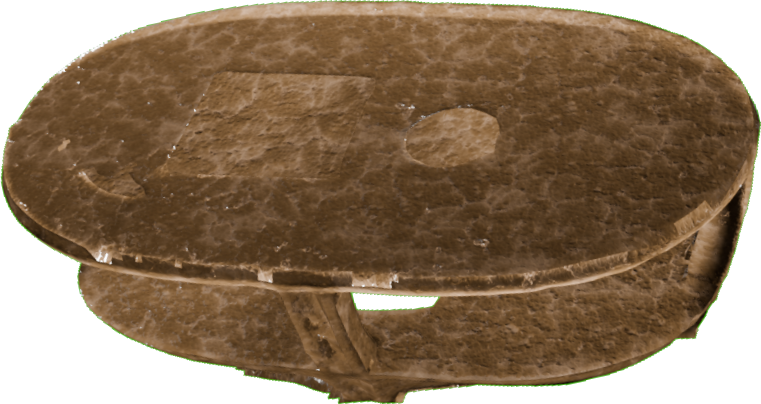}
      \caption{Leather}
    \end{subfigure}
    &
    \begin{subfigure}[b]{0.2\textwidth}
      \centering
      \includegraphics[width=\textwidth]{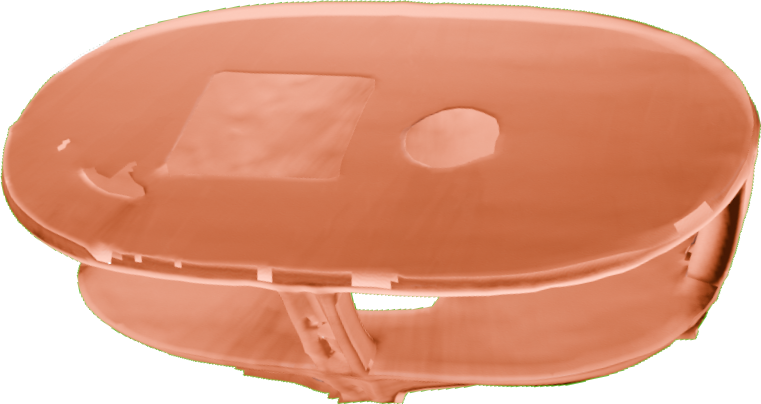}
      \caption{Plastic}
    \end{subfigure}
    &
    \begin{subfigure}[b]{0.2\textwidth}
      \centering
      \includegraphics[width=\textwidth]{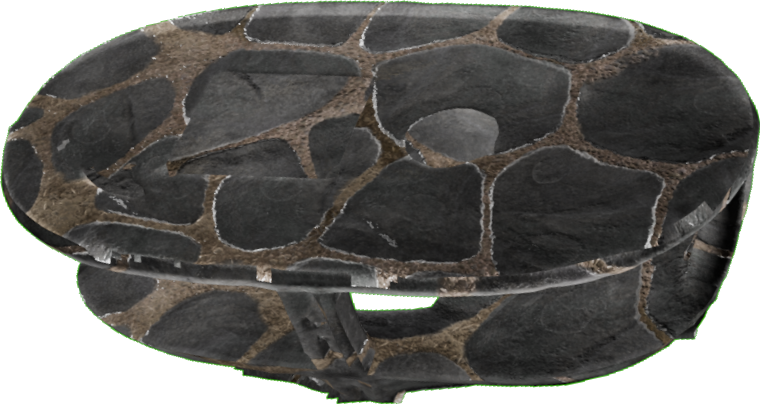}
      \caption{Stone}
      \label{fig:bmp}
    \end{subfigure}
  \end{tabular}
  \caption{Table rendered by different materials}
  \vspace{-10pt}
  \label{fig:table-rendering}
\end{figure}

Inverse rendering aims to estimate physical attributes of a scene from images. While there have been multiple approaches \cite{li2018cgintrinsics, sengupta2019neural} to estimate inverse rendering, it remains a complex problem due to interplay between appearance of different objects - occlusions and shadows which can change appearance. Another aspect which makes the problem challenging is the material diversity of a common object - which none of the previous research talks about. Specifically, a single indoor object could be made of different types of material. For example, a table could be made of wood, stone, plastic etc. while still retaining the same shape (Fig. \ref{fig:table-rendering}). 

We curate a large scale dataset aimed towards creating different versions of the same object based on different material properties. We utilize Replica, introduced by Meta, which is a dataset of 18 highly photo-realistic 3D indoor scene reconstructions \cite{straub2019replica}. Each scene in the dataset consists of a dense mesh, high-resolution high-dynamic-range (HDR) textures, per-primitive semantic class and instance information and reflective surfaces. This makes the dataset very realistic and a much better alternative to synthetic datasets such as SUNCG \cite{song2017semantic} in terms of semantic richness. For each scene, we isolate each object sub-mesh separately along with the HDR textures from the global scene mesh. In order to output different material properties, we rely on MatSynth \cite{vecchio2024matsynth} dataset to query specific material properties. MatSynth contains from than 4000 CC0 ultra-high resolution PBR materials. We then import the specific object mesh into Blender \cite{Blender2018modelling} and insert the material properties associated with different materials through the Principled BSDF shader. This generates the object with the target material. We then reinsert the new perturbed object into the scene mesh. The camera placement is made by dividing the spherical volume around the object into a grid and then placing the camera on the grid. We also aim to provide a benchmark for inferring the material properties given these perturbed objects. For a given image in the scene, we isolate each object using a semantic segmentation step, using segment anything (SAM) \cite{kirillov2023segment}. We then insert this object into some chosen neural network architectures in order to infer the basecolor property. We show that these architectures are able to recover these material properties in diverse conditions. 

Our goal through this dataset and benchmark is to help the computer vision community working on indoor robotics applications. Indoor environments are extremely rich and varied and navigating these environments autonomously presents a immense challenge. Adding to that, the robots need to perform chores (``get the glass jug of water from the counter while not disturbing others") and we can see why indoor robotics is progressing at a much slower pace. Our efforts are in that general direction and we aim to better reflect the richness observed in the indoor environment. 
We provide some literature review regarding the datasets and methods for inverse rendering in Section \ref{sec:background}. In Section \ref{sec:dataset}, we explain the dataset in detail and in Section \ref{sec:benchmark}, we provide the benchmark including the different architectures compared, evaluation metrics and results. Finally we conclude with some discussions and future work in Section \ref{sec:conclusions}.  

\section{Background}
\label{sec:background}
As discussed above, material properties is a composite of various different properties. Primary among these is texture analysis, which has long been a fundamental and challenging problem in pattern recognition requiring classification, segmentation, synthesis and shape from texture \cite{tuceryan1993texture}. Traditional pattern recognition methods were proposed for texture identification such as Bag of Words (BoW) with a universal dictionary for learning textures of all images \cite{leung2001representing}. 
With the advent of Convolutional neural networks (CNN), several CNN variants have been proposed for material recognition with texture specific ScatNet \cite{bruna2013invariant} and PcaNet \cite{chan2015pcanet} outperforming other deep learning based methods. One roadblock with deep learning methods is the requirement of large datasets of images with domain-specific material properties and several representative samples in each category captured under different illumination and viewing conditions \cite{tremeau2020deep}. This is an ill-posed and under-constrained problem without a general solution. Usually, there are either implicit or explicit methods where image-based representations are used to interpolate novel views or simulation-based representations which extrapolate new views from simulation. 
One possible alternative is to learn view-independent appearance features (or shape-independent appearance features) \cite{arto2019similarity}. However, utilizing a crowd-sourced measure of similarity as described in the paper is neither robust nor scalable. 
Therefore, we need to look at a radically different approach to produce a wealth of images on command under different viewing and illumination conditions. 

The SUNCG dataset \cite{song2017semantic} provided the first example of high-quality synthetic dataset essential for inverse rendering. However, their datasets were rendered with OpenGL under fixed point light sources. The PBRS dataset \cite{zhang2017physically} extended the SUNCG dataset by using PBR with a physics based renderer called Mitsuba \cite{jakob2022mitsuba3}. The rendered images were very noisy, all materials were treated as diffused and a single outdoor environment map. CGIntrinsics \cite{li2018cgintrinsics} modified PBRS using the computationally expensive Bidirectional path tracing (BDPT). On the other hand, CG-PBR \cite{sengupta2019neural} modified the SUNCG by rendering under multiple outdoor environment maps and rendering the same scene twice, using Lambertian surfaces and with default settings. While these research works point to learning of different material properties, they are very constrained with respect to objects, materials and illumination. 

\section{The MatPredict Dataset}
\label{sec:dataset}
While the previous datasets show that predicting material properties from camera images is possible, they do not address the heterogeneity of the problem space. We therefore address the question of material diversity by generating a synthetic dataset with different material properties. 
We can then insert the Replica dataset \cite{straub2019replica} into the simulation, an open-source dataset released by Meta, of high-quality reconstructions of various indoor environments along with glass surfaces and textures information of objects present in the scene. We would like to change the material properties of the objects in the scene. To do this, we utilize MatSynth \cite{vecchio2024matsynth} dataset, which consists of different material categories and their properties. Given this combination, we can generate a large distribution of realistic objects, composed of different materials, in the indoor environment. Below we detail the steps taken towards generating the dataset in detail.

\paragraph{Mesh file separation and texture rendering}
We begin by extracting a global mesh of each indoor scene from the Replica dataset\cite{straub2019replica} and use the per-face instance IDs to partition this mesh into individual object sub-meshes. For material appearance, we query the MatSynth \cite{vecchio2024matsynth} dataset and retrieve the calibrated texture bundle associated with every material class—namely basecolor, diffuse, metallic, normal, opacity, and roughness maps. Each sub-mesh is then paired with the texture set that corresponds to its semantic material label. The textured sub-meshes are imported into Blender, where a procedural node graph feeds the texture maps into a Principled BSDF shader during Cycles rendering.

\begin{figure*}[t]
  \centering
  \subfloat[$s=0.1$]{%
    \includegraphics[width=0.19\linewidth]{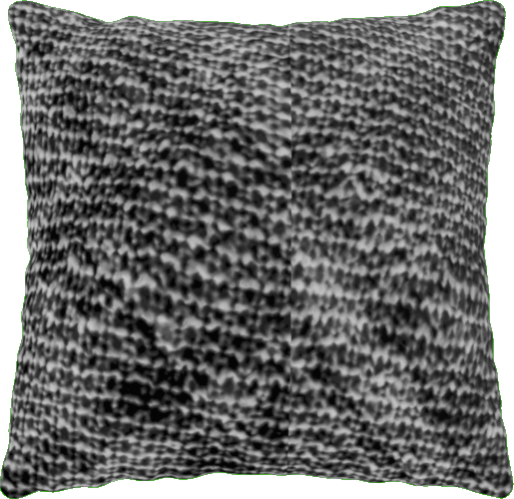}}\hfill
  \subfloat[$s=0.2$]{%
    \includegraphics[width=0.19\linewidth]{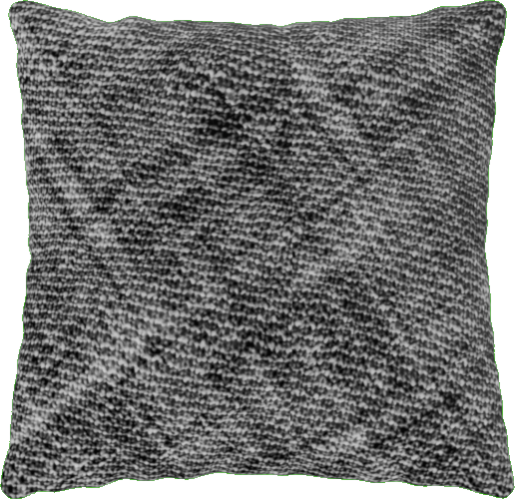}}\hfill
  \subfloat[$s=0.5$]{%
    \includegraphics[width=0.19\linewidth]{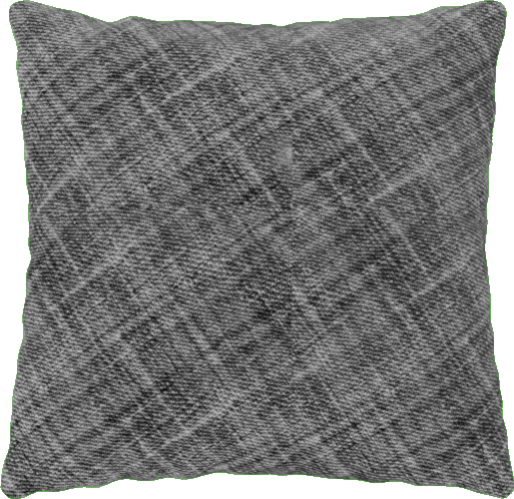}}\hfill
  \subfloat[$s=1.0$]{%
    \includegraphics[width=0.19\linewidth]{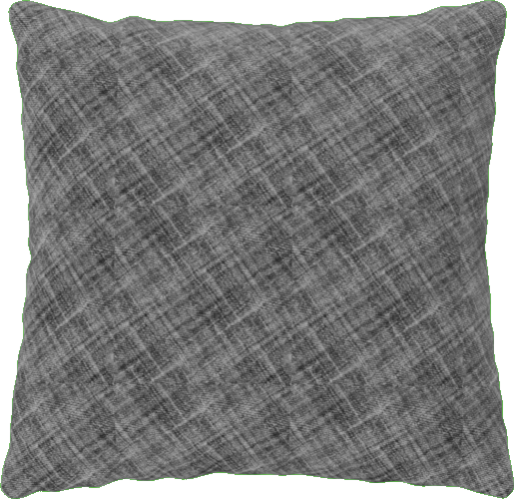}}\hfill
  \subfloat[$s=2.0$]{%
    \includegraphics[width=0.19\linewidth]{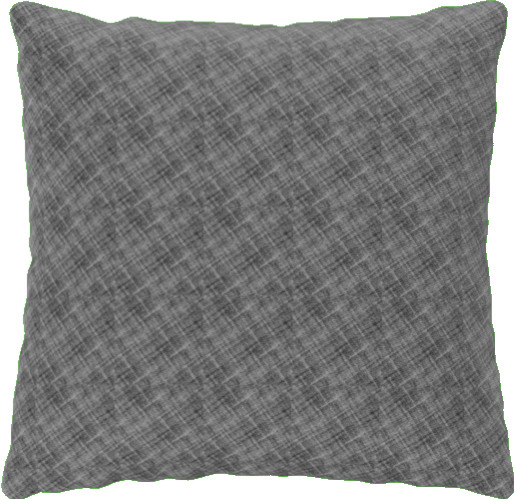}}
  \caption{Effect of the UV scale factor \(s\) on appearance.  
           Smaller \(s\) values map a larger texture footprint to the surface, producing finer weave patterns, whereas larger \(s\) values stretch the same fabric map, leading to visibly coarser detail.}
  \label{fig:uv-scale-study}
\end{figure*}

\paragraph{UV texture preprocessing.}
Before any rendered frame enters our pipeline we perform a single, automated UV-editing pass that prepares every sub-mesh for texture look-up.  
First, a \emph{texel-density normalisation} stage rescales the UV islands so that the physical texel \cite{RTR4} pitch is consistent across objects of very different size: given a surface area \(A\) we estimate a characteristic length \(\ell=\sqrt{A/\pi}\) and set the target texel density to \(d = d_{0}\,(\ell/\ell_{0})^{-1}\) with \(d_{0}=512\,\text{px\,m}^{-1}\) at \(\ell_{0}=1\,\text{m}\).  
Because all MatSynth material maps share the same native resolution, this adaptive scaling guarantees that a \(4\text{k}\) texture represents comparable real-world detail—whether it is applied to a teacup or a wardrobe. Fig. \ref{fig:uv-scale-study} illustrates the visual impact of the scale factor with five renderings of the same \textit{fabric} pillow at \(s\in\{0.1,0.2,0.5,1.0,2.0\}\).

Next, the surface is partitioned along high-curvature ridges and discontinuous normals into near-planar charts that can be unfolded with minimal angular distortion.  Each chart is flattened by an angle-based conformal algorithm (conceptually similar to “smart UV” in Digital Content Creation (DDC) suites \cite{BlenderSmartUV}), which equalizes stretch while preserving edge adjacency.  The resulting charts are then greedily packed into sequential U-Dimension (UDIM) tiles with a fixed two-pixel gutter; this both maximizes texture utilization and stops colors from leaking into neighboring UV islands when the texture is shrunk to low-resolution mipmaps \cite{Williams1983Mipmaps}.

Together, density normalization, scale adaptation and distortion-controlled chart packing yield seam-free UV layouts whose texel resolution is physically meaningful and uniform throughout the dataset.

\paragraph{Camera placement.}
For every target object we generate a deterministic, latitude--longitude grid on a spherical shell of fixed radius \(r\) centred at the object’s geometric centroid \(\mathbf{c}\).
Let
\[
  \phi\in[\phi_{\min},\phi_{\max}],\qquad
  \theta\in[\theta_{\min},\theta_{\max}]
\]
denote the polar (elevation) and azimuth angles, respectively.
The default configuration uses
\(\phi_{\min}=0,\;\phi_{\max}=\tfrac{\pi}{2}\) and
\(\theta_{\min}=-\tfrac{\pi}{2},\;\theta_{\max}=\tfrac{\pi}{2}\);
i.e.\ the camera moves over the front hemi-sphere that faces the viewer.
Any sub-range can be specified at run time to tailor the coverage to elongated or asymmetric objects.

\begin{figure*}[t]
  \centering
  \subfloat[$(15^{\circ},-60^{\circ})$]{%
    \includegraphics[width=0.15\linewidth]{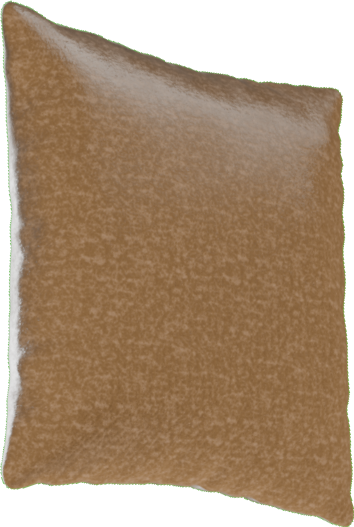}}\hfill
  \subfloat[$(15^{\circ},\;0^{\circ})$]{%
    \includegraphics[width=0.15\linewidth]{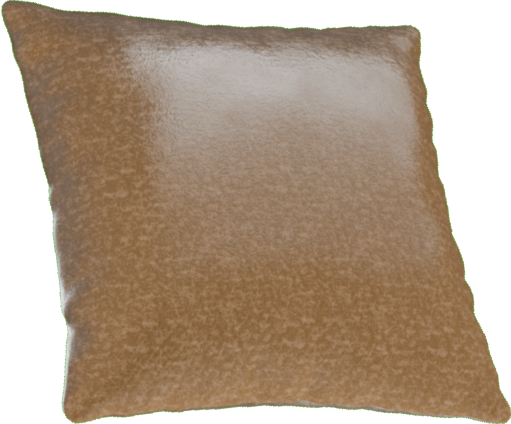}}\hfill
  \subfloat[$(15^{\circ},\;60^{\circ})$]{%
    \includegraphics[width=0.15\linewidth]{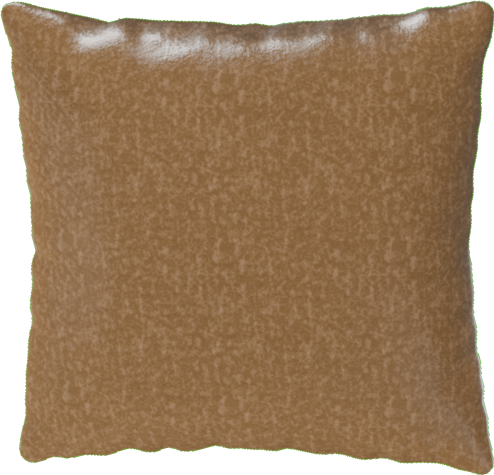}}\hfill
  \subfloat[$(45^{\circ},-30^{\circ})$]{%
    \includegraphics[width=0.15\linewidth]{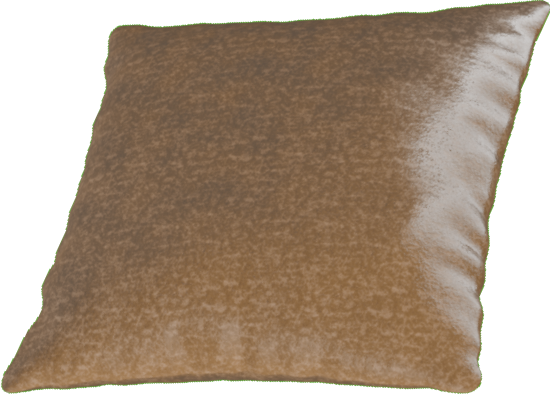}}\hfill
  \subfloat[$(45^{\circ},\;30^{\circ})$]{%
    \includegraphics[width=0.15\linewidth]
    {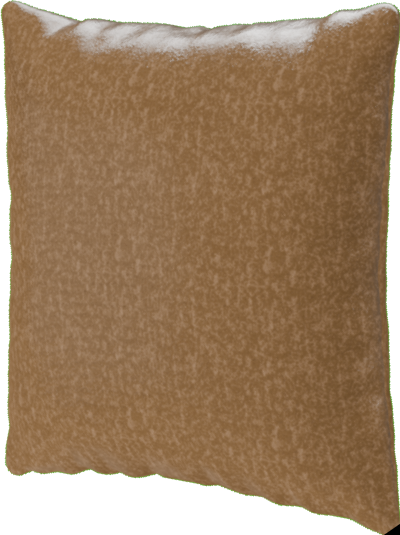}}\hfill
  \caption{Effect of viewpoint on appearance.  
           Five leather-pillow renderings sampled from the latitude–longitude grid
           demonstrate how changes in elevation~\(\phi\) and azimuth~\(\theta\)
           influence specular highlights, perceived shape, and shadow placement.}
  \label{fig:view-study}
\end{figure*}

With \(N_\phi\) latitudinal and \(N_\theta\) longitudinal divisions, the cell centres
\begin{equation}
  \phi_i=\phi_{\min}+\Bigl(i+\tfrac12\Bigr)\frac{\phi_{\max}-\phi_{\min}}{N_\phi},
  \qquad
  \theta_j=\theta_{\min}+\Bigl(j+\tfrac12\Bigr)\frac{\theta_{\max}-\theta_{\min}}{N_\theta}
\end{equation}
define \(N_\phi\times N_\theta\) camera positions
\begin{equation}
  \mathbf{p}_{ij}= \mathbf{c}+r\,
  \bigl[\sin\phi_i\cos\theta_j,\;
        \sin\phi_i\sin\theta_j,\;
        \cos\phi_i\bigr]^{\!\top}.
\end{equation}
Each camera is then rotated so that its optical \texttt{–Z} axis points exactly to \(\mathbf{c}\) and its \texttt{Y} axis remains vertical, yielding upright images irrespective of viewpoint.
In practice we set \(N_\phi=16,\;N_\theta=32\) and \(r=1.0\;\mathrm{m}\), producing \(512\) uniformly stratified viewpoints per object; adjusting \(N_\phi, N_\theta\) or shrinking \([\phi_{\min},\phi_{\max}], [\theta_{\min},\theta_{\max}]\) immediately refines or sparsifies the capture without altering the pipeline. Figure~\ref{fig:view-study} visualises the resulting variation with five renderings of a leather pillow captured from representative grid positions.

\paragraph{Lighting rig and source parameters.}
To obtain uniform and reproducible illumination, we surround every object with a symmetric set of low-power lamps.
Given the radius \(r\) and centre \(\mathbf{c}\) of the object’s minimal enclosing sphere, we place a \emph{key--fill ring} of
\begin{equation}
  N_\theta=\bigl\lceil4+2\,r/0.25\bigr\rceil,\qquad 6\le N_\theta\le 12
\end{equation}
rectangular \emph{area} lights on the horizontal circle of radius \(2r\) through~\(\mathbf{c}\).
Each lamp faces~\(\mathbf{c}\); its long edge is tangent to the circle and its size is fixed to \(0.6r\times0.3r\), so neighbouring penumbras overlap smoothly.
Two additional accent lights are added on the vertical axis at~\(\pm35^{\circ}\).
If \(r>0.15\,\mathrm{m}\) these accents are area lights; otherwise they are \(10^{\circ}\) spot lights, both aimed at~\(\mathbf{c}\).

The total radiant flux is distributed with a cosine fall--off, keeping the illuminance at~\(\mathbf{c}\) at \(E\approx1.0\,\mathrm{k\,lx}\,(\pm5\%)\).
The power of an individual lamp lies in the \textbf{20--200~W} range and scales with object size:
\begin{equation}
    P_{ij}=P_\text{base}\bigl(1+0.3\cos\theta_j\bigr),\qquad
  P_\text{base}=50+150\min\bigl(1,\,r/1\,\mathrm{m}\bigr).
\end{equation}

\begin{figure*}[t]
  \centering
  \subfloat[1~W]{\includegraphics[width=0.15\linewidth]{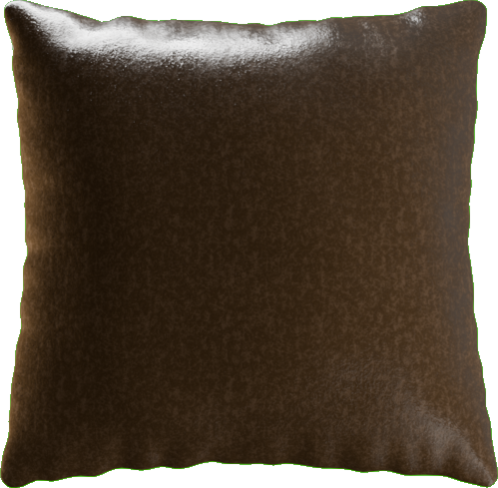}}\hfill
  \subfloat[10~W]{\includegraphics[width=0.15\linewidth]{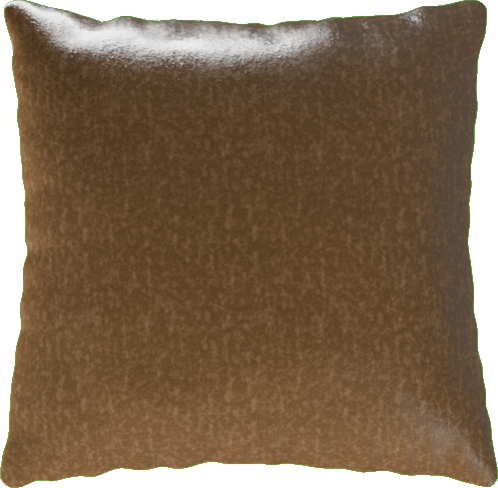}}\hfill
  \subfloat[40~W]{\includegraphics[width=0.15\linewidth]{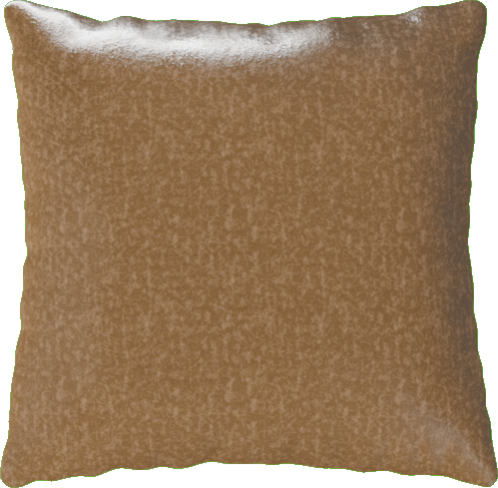}}\hfill
  \subfloat[200~W]{\includegraphics[width=0.15\linewidth]{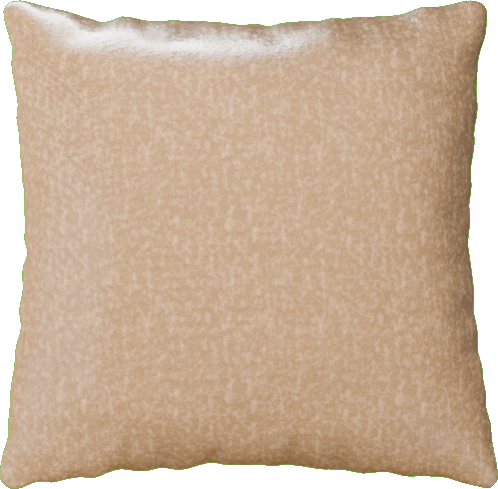}}\hfill
  \subfloat[1000~W]{\includegraphics[width=0.15\linewidth]{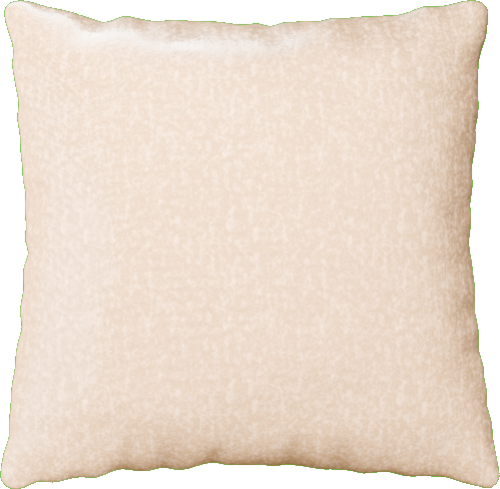}}
  \caption{leather pillow with different lighting setup}
  \vspace{-5pt}
  \label{fig:intensity-study}
\end{figure*}

To avoid colour casts every emitter uses the neutral grey--white Blender RGB value \textbf{(0.8,\,0.8,\,0.8)}, corresponding to a correlated colour temperature of about \(5{,}800\,\mathrm{K}\).
This multi-source configuration keeps the object at the photometric centre of the scene, suppresses deep cast shadows and excessive inter-reflections, and removes the need for manual tweaking.
Figure~\ref{fig:intensity-study} shows five renderings of a leather pillow under progressively stronger lamp powers, illustrating the influence of the 1~W\,→\,1000~W range on appearance.

\section{Benchmark}
\label{sec:benchmark}

\subsection{Image-to-Basecolor Prediction Pipeline}
\label{sec:pipeline}
Given a cropped RGB frame \(\mathbf{I}\in\mathbb{R}^{3\times224\times224}\)
rendered by the \textsc{MatPredict} simulator—each originating from one of the
object–material pairs listed in Table~\ref{tab:input_table}—our goal is to
recover the pixel-wise \emph{basecolour}
\(\mathbf{\hat B}\in\mathbb{R}^{3\times224\times224}\).
Each crop is normalized, optionally center- or random-cropped, and passed through
a neural network \(f_\theta\) that comprises an \textbf{encoder} \(E_\theta\) and a
\textbf{decoder} \(D_\theta\):
\begin{equation}
      \mathbf{\hat B}=f_\theta(\mathbf I)
  =D_\theta\!\bigl(E_\theta(\mathbf I)\bigr).
\end{equation}

Networks are trained end-to-end with an \(\mathcal{L}_\text{MSE}\) loss, the Adam
family of optimizers, and an identical learning-rate schedule across all
experiments (experiments details in Table~\ref{tab:mixed_table} and Appendix Table ~\ref{tab:mixed_table_roughness} ).

\subsection{Architectures evaluated}
\label{sec:arch}
To quantify task difficulty across architectural families we benchmark four
encoders of increasing sophistication (Table \ref{tab:arch}):
\begin{enumerate}[label=(\arabic*)]
\item a compact \textsc{UNet} without long skips,
\item \textsc{ResNet-50} with a lightweight decoder,
\item the Transformer-based \textsc{Swin-T}, and
\item \textsc{ConvNeXt-Tiny}, a modern CNN that bridges convolutional efficiency
      with Transformer-style macro design.
\end{enumerate}
All networks ingest \(224^{2}\) crops and output equally sized basecolor maps,
enabling a strict apples-to-apples comparison.

\begin{table}[h]
  \centering
  \caption{Network backbones evaluated in our benchmark.
           All variants output a $224\times224$ RGB basecolor map.}
  \label{tab:arch}
  \begin{tabular}{lccc}
    \toprule
    Model & Encoder backbone & Decoder scheme & \#Params \\
    \midrule
    UNet-no-skip \cite{Ronneberger2015UNet} & conv\,$\times4$ (64–512) & 4-stage upconv, no long skip & 23.6M \\
    ResNet50-U \cite{He2016ResNet}   & ImageNet ResNet-50       & 5-stage upconv (512\(\rightarrow\)16) & 39.5 M \\
    Swin-T \cite{Liu2021Swin}      & Swin Tiny, patch4        & 5-stage upconv (384\(\rightarrow\)24) & 31.6 M \\
    ConvNeXt-T\cite {Liu2022ConvNeXt}    & ConvNeXt Tiny             & 5-stage upconv (384\(\rightarrow\)24) & 32.0 M \\
    \bottomrule
  \end{tabular}
\end{table}

\subsection{Training protocol}
\label{sec:train}

\paragraph{Dataset preparation.}
For every material class \(\langle m\rangle\in\mathcal{M}\) and object
category \(\langle o\rangle\in\mathcal{O}\) the simulator exports  

\begin{itemize}[nosep,leftmargin=*]
  \item \(512\) RGB screenshots  
        \texttt{rendered\_cropped/\(\langle o\rangle\)/\(\langle m\rangle\)/*.png},
  \item one reference \emph{basecolour} map  
        \texttt{ground\_truth\_basecolour/\(\langle m\rangle\).png},
  \item one reference \emph{roughness} map  
        \texttt{ground\_truth\_roughness/\(\langle m\rangle\).png}.
\end{itemize}

During initialisation the PairedImageDataset
 
\begin{enumerate}[label=(\alph*)]
  \item loads both \(\mathbf{B}_m\) and \(\mathbf{R}_m\) into RAM once,
  \item enumerates screenshot indices
        \(\langle\mathbf{I}_{m,o,k},\,m\rangle\),
  \item and stores them in a flat list \(\mathcal{S}\)
        of length \(|\mathcal{M}|\!\times\!|\mathcal{O}|\!\times512\).
\end{enumerate}
With a fixed seed (\texttt{42}) we shuffle \(\mathcal{S}\) once and split
it 80 
\textsc{Subset}.

\paragraph{Targets and loss.}
For each sample we stack the basecolour and roughness maps channel-wise
to obtain a 6-channel target  
\(\mathbf{T}=\bigl[\mathbf{B}\,\|\,\mathbf{R}\bigr]
 \in\mathbb{R}^{6\times224\times224}\).
All decoders therefore end with a \(3\!\rightarrow\!6\) \(1\times1\) conv.
The training objective is an equally weighted sum of two MSE terms:
\begin{equation} 
  \mathcal{L}
  =\underbrace{\|\hat{\mathbf{B}}-\mathbf{B}\|_2^2}_{\mathcal{L}_{\text{B}}}
  +\underbrace{\|\hat{\mathbf{R}}-\mathbf{R}\|_2^2}_{\mathcal{L}_{\text{R}}}.
\end{equation}

\paragraph{Pre-processing.}
Screenshots are resized to \(224\times224\).
Swin-T inputs are normalised with ImageNet mean/std, whereas the other
backbones consume raw \([0,1]\) tensors.
We keep \texttt{--if\_cropped False} so that input and both targets are
pixel-aligned.

\paragraph{Optimisation details.}
Unless stated otherwise we train for 50 epochs with a batch size of 8
(\texttt{num\_workers=4}).  

\begin{itemize}[nosep,leftmargin=*]
  \item \textbf{UNet, ResNet-50, ConvNeXt-T}: Adam optimiser, initial 
        learning rate \(\eta_{0}=2\times10^{-4}\).
  \item \textbf{Swin-T}: AdamW optimiser,
        \(\eta_{0}=1\times10^{-4}\), weight decay \(10^{-2}\).
  \item \textbf{LR schedule}: StepLR
        (\texttt{step\_size}=20, \(\gamma=0.5\); default); 
        CosineAnnealing (\(T_{\max}=50\)) or ReduceLROnPlateau
        selectable via \texttt{--learning\_rate\_schedule}.
\end{itemize}

\paragraph{Device.}
All experiments are executed on a desktop workstation with a single
NVIDIA GeForce RTX 4070 Super GPU.


\subsection{Extensibility to additional material channels}
\label{sec:extensibility}

A key advantage of the encoder–decoder back-bone used in all four
benchmarks is that \emph{the only component that depends on the number of
predicted material layers is the final \(1{\times}1\) convolution}.  In
§\ref{sec:train} we demonstrated a two-head variant that jointly regresses
basecolour \(\mathbf{B}\) and roughness \(\mathbf{R}\)
(\(C{=}6\) output channels) (results for roughness in Table \ref{tab:mixed_table_roughness}).  Generalizing to further physical attributes
is therefore straightforward:

\begin{itemize}[nosep,leftmargin=*]
  \item \textbf{Head adaptation} —  
        replace the last convolution by one with
        \(C = 3{+}N_{\text{extra}}\) kernels, e.g.\ +1 for metallicity
        or +3 for a normal map in \((x,y,z)\).
  \item \textbf{Loss formulation} —  
        form a channel-wise stack
        \(\mathbf{T}=[\mathbf{B}\,\|\,\mathbf{R}\,\|\,\mathbf{M}\,\|\,\mathbf{N}]\)
        and minimise
        \begin{equation}
          \mathcal{L}
          =\sum_{c=1}^{C}w_{c}\,
            \bigl\|\hat{\mathbf{T}}_{c}-\mathbf{T}_{c}\bigr\|_{2}^{2},
        \end{equation}
        where \(w_{c}\) can be used to balance heterogeneous ranges
        across layers (e.g.\ metallic vs.\ colour).
  \item \textbf{Training protocol} —  
        no other hyper-parameter changes are required; batch size,
        optimiser, and learning-rate schedule transfer unchanged.
\end{itemize}

\subsection{Image–image evaluation metrics}
\label{sec:metrics}

The eight metrics used in this study fall naturally into three
families—\emph{error–ratio}, \emph{perceptual / structural},
and \emph{spectral / information–theoretic}.
Table~\ref{tab:mixed_table} in the main paper reports one
representative metric per family, while the complete mathematical
definitions of all eight indices can be found in
App.~\ref{app:metrics}. Table~\ref{tab:metric-formulas}.
gives a concise, interpretation–oriented overview.

\begin{table}[t]
  \centering
  \caption{Eight image-similarity metrics grouped by family.  
           Unless noted otherwise, larger values indicate better
           similarity; RMSE and SAM are inverse metrics where lower is
           better.}
  \label{tab:metric-summary}
  \setlength{\tabcolsep}{4pt}
  \renewcommand{\arraystretch}{1.15}
  \begin{tabular}{llp{6.5cm}l}
    \toprule
    Family & Metric & Brief description & Range / direction \\
    \midrule
    \multirow{2}{2cm}{Error–ratio}%
      & RMSE  & Root mean-square pixel error. & $[0,\infty)$, lower is better \\
      & PSNR  & Peak signal-to-noise ratio in dB. & $[0,\infty)$, higher is better \\
      & SRE   & Signal-to-reconstruction error (dB), PSNR w.r.t.\ image energy. & $[0,\infty)$, higher is better \\[2pt]
    \multirow{2}{2cm}{Perceptual / structural}%
      & SSIM  & Structural similarity (luminance, contrast, structure). & $[-1,1]$, higher is better \\
      & FSIM  & Feature similarity based on phase congruency and gradient magnitude. & $[0,1]$, higher is better \\
      & UIQ   & Universal image quality index (combined luminance/contrast/structure). & $[-1,1]$, higher is better \\[2pt]
    \multirow{2}{2cm}{Spectral / info-theoretic}%
      & SAM   & Spectral angle mapper—mean angular error in colour space. & $[0,90^{\circ}]$, lower is better \\
      & ISSM  & Information-theoretic statistic similarity (relative Frobenius norm). & $(0,1]$, higher is better \\
    \bottomrule
  \end{tabular}
\end{table}

\begin{table}[ht]
  \centering
  \caption{Different objects rendered with the chosen materials (partial)}
  \setlength{\tabcolsep}{1pt}
  \renewcommand{\arraystretch}{1.2}

  \begin{tabularx}{1\textwidth}{>{\raggedright}p{1.2cm} p{2cm} p{2cm} p{2cm} p{2 cm}p{2 cm} p{2.0cm} p{2.0cm}p{2cm}p{1.2cm}}
    \toprule
       \textbf{Material} & \textbf{Table} & \textbf{Chair} & \textbf{Pillow} & \textbf{Cabinet} & \textbf{Vase} & \textbf{Sofa}  \\ \midrule
    Wood    & \imgcell{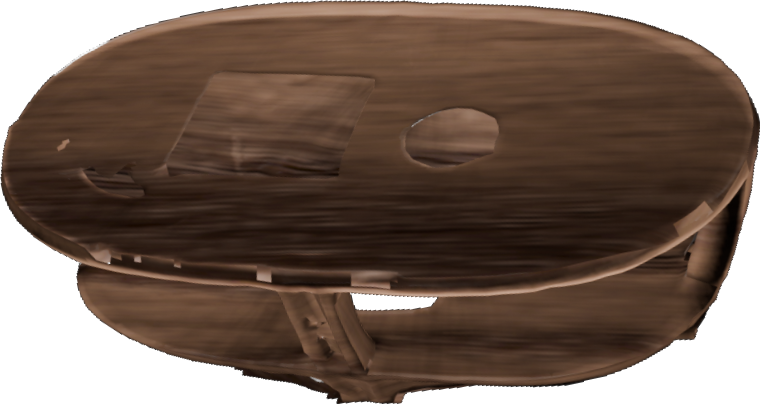}
            & \imgcell{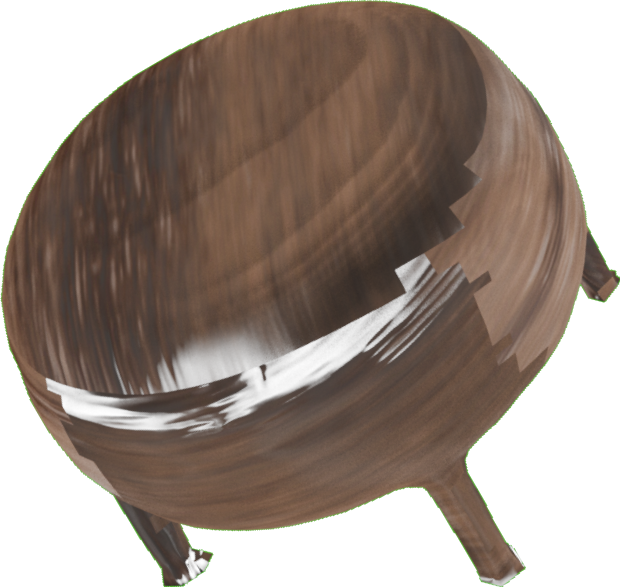}  
            & \imgcell{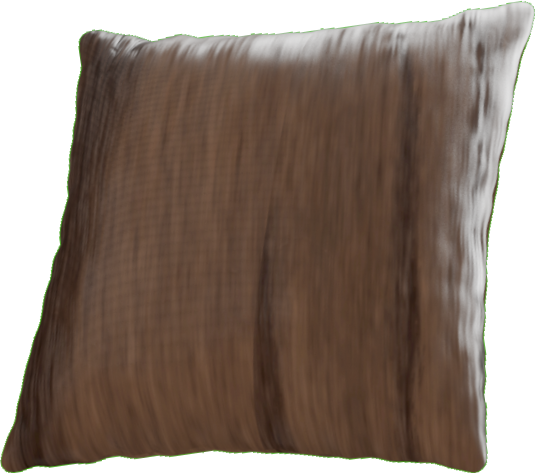} 
            & \imgcell{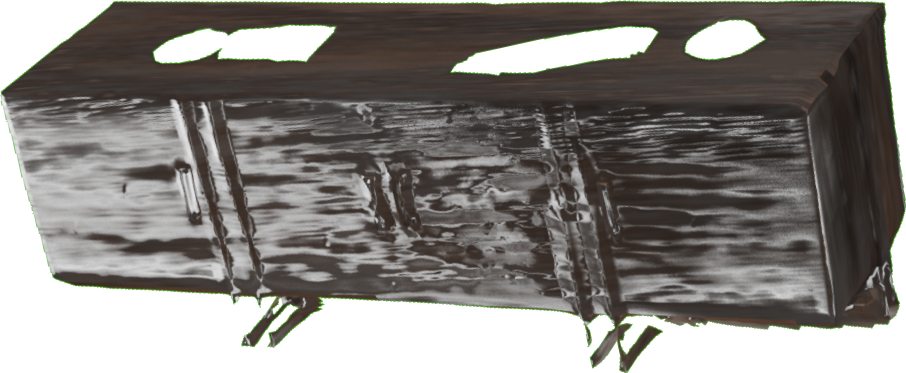} 
            & \imgcell{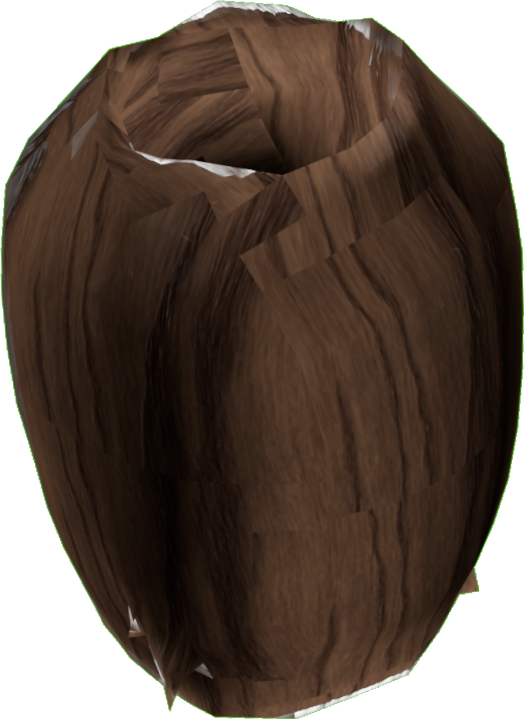} 
            & \imgcell{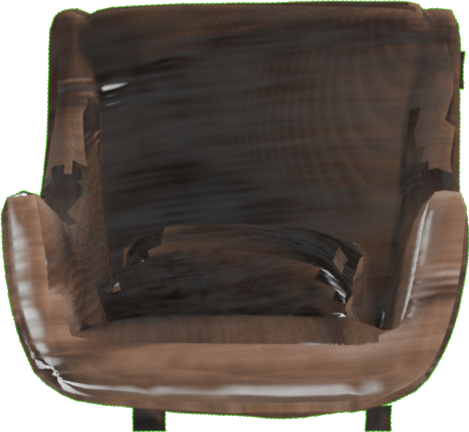} 
            \\
    Concrete 
            & \imgcell{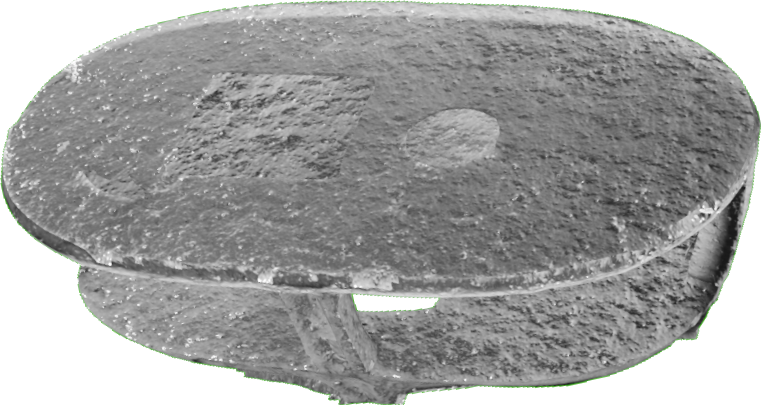}
            & \imgcell{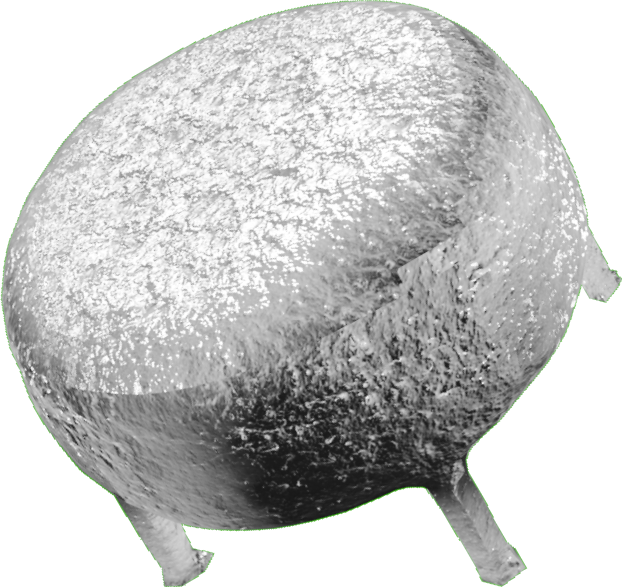}  
            & \imgcell{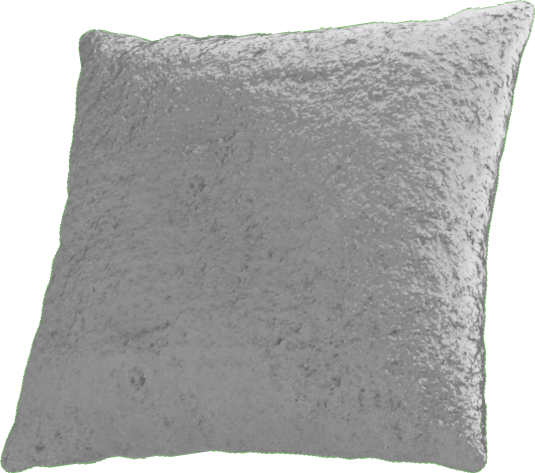} 
            & \imgcell{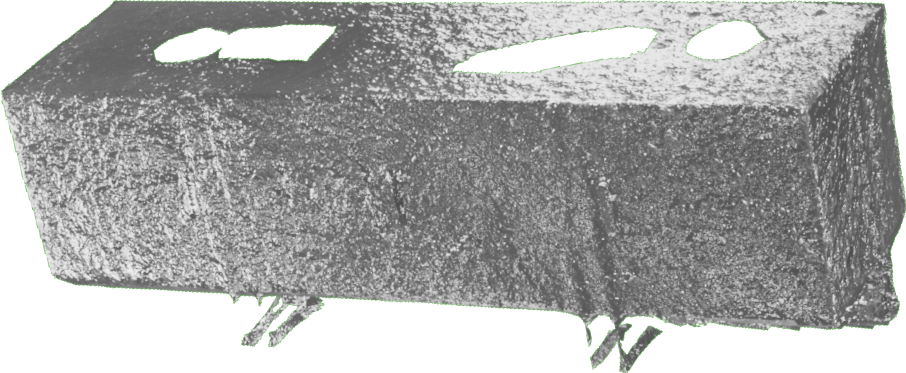} 
            & \imgcell{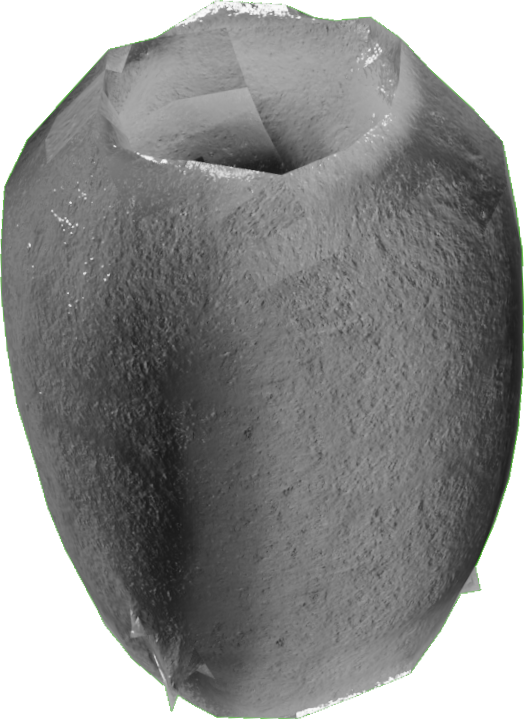} 
            & \imgcell{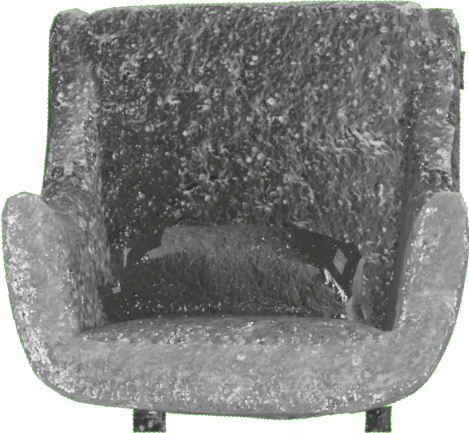} 
         \\
    Plastic 
            & \imgcell{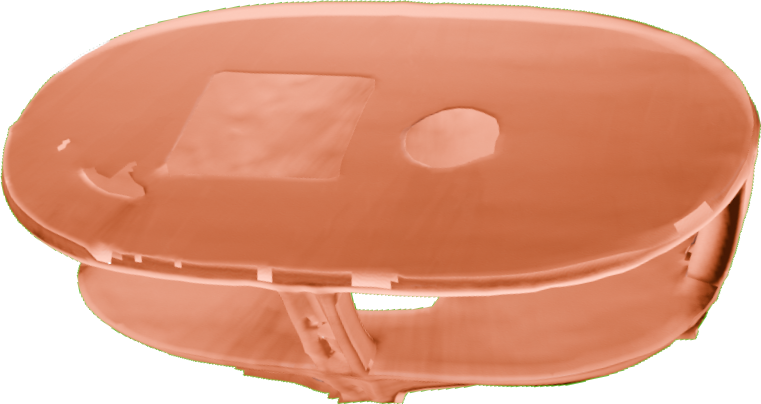}
            & \imgcell{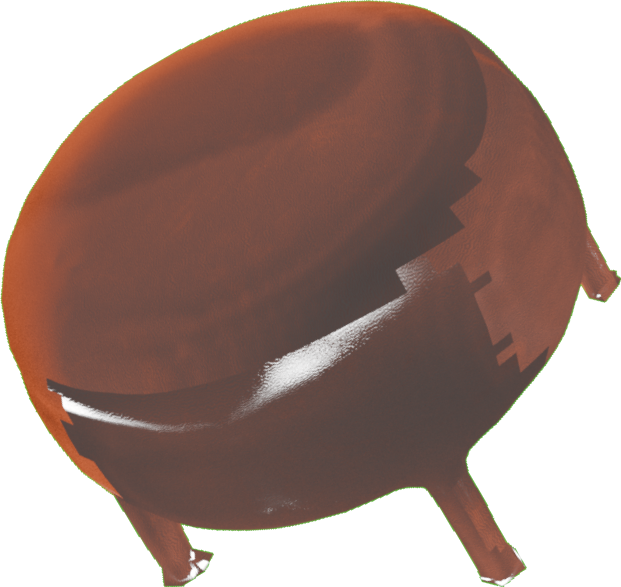}  
            & \imgcell{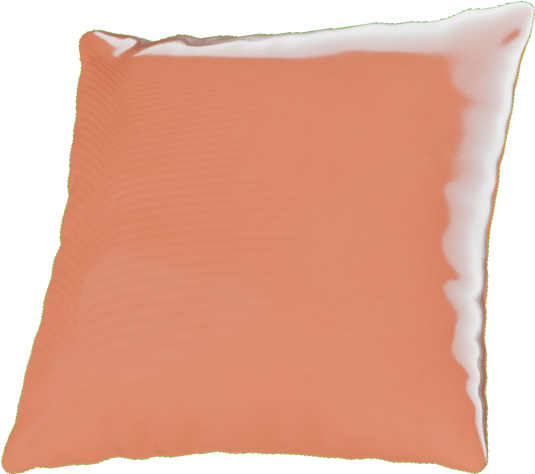} 
            & \imgcell{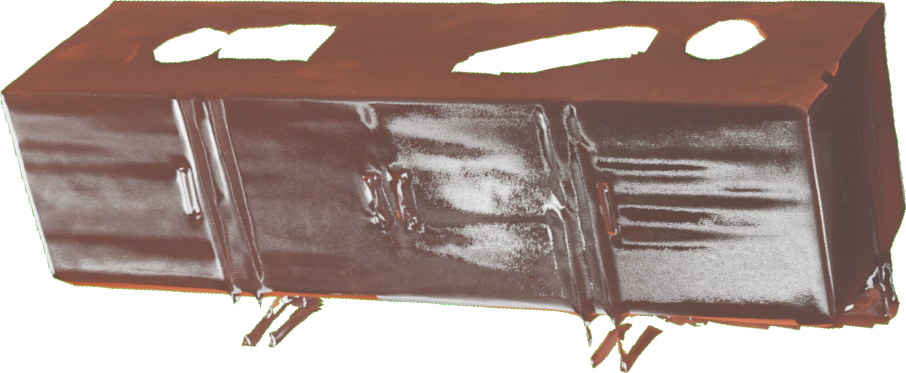} 
            & \imgcell{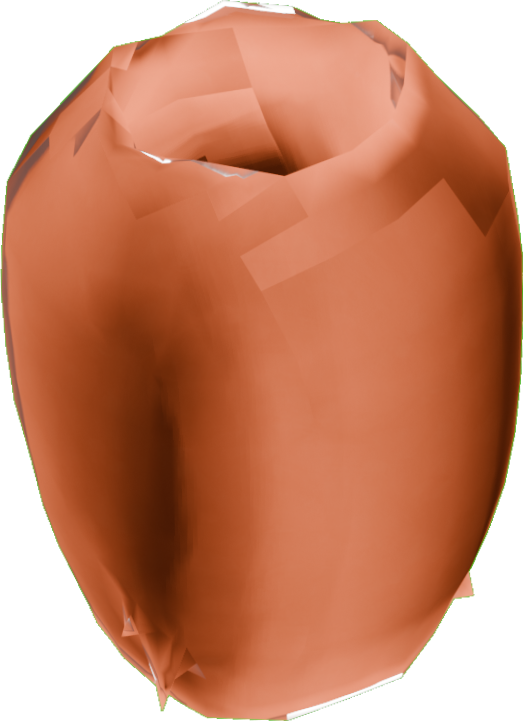} 
            & \imgcell{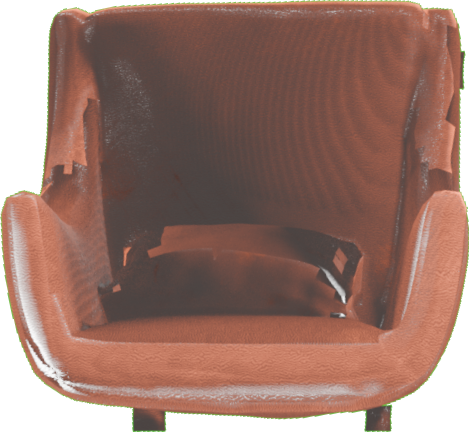} 
        \\
    Stone  
            & \imgcell{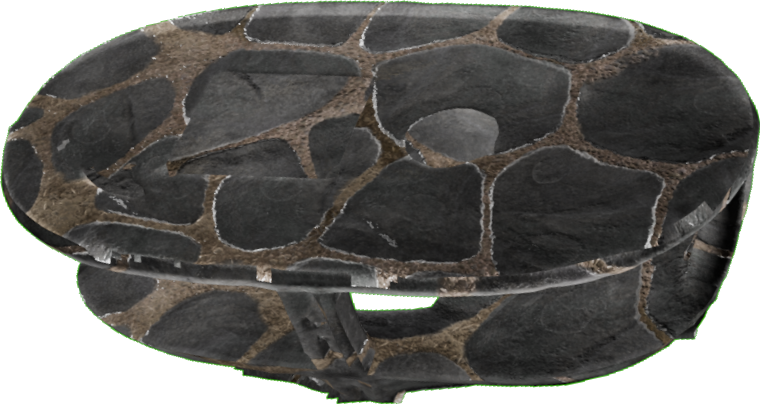}
            & \imgcell{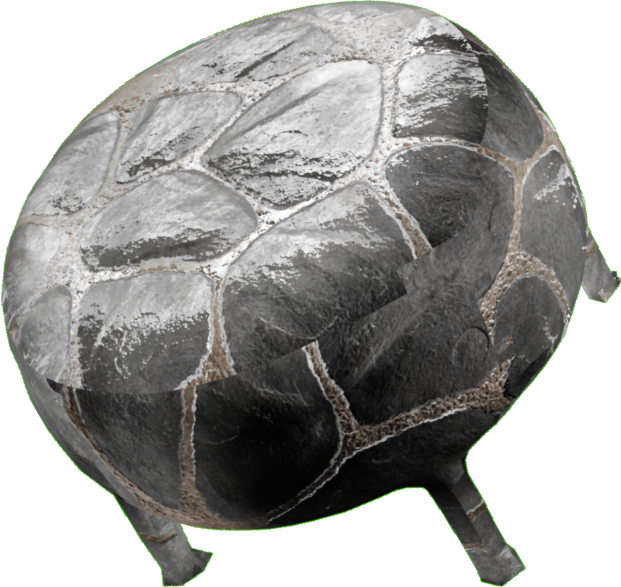}  
            & \imgcell{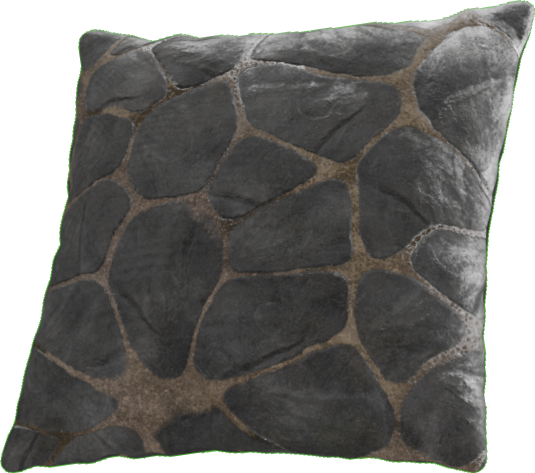} 
            & \imgcell{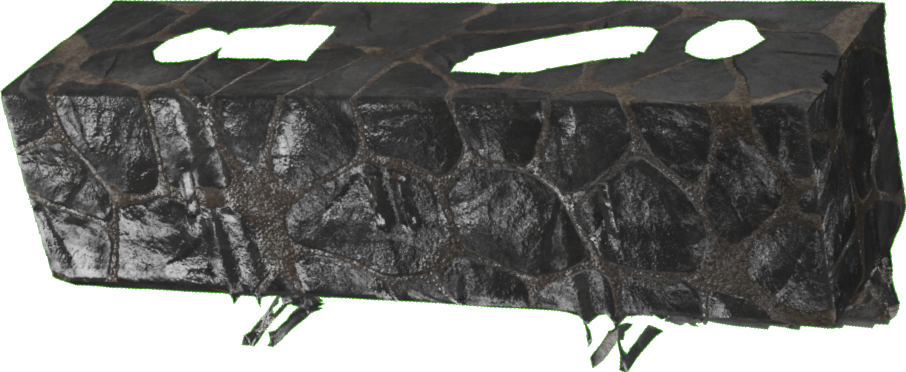} 
            & \imgcell{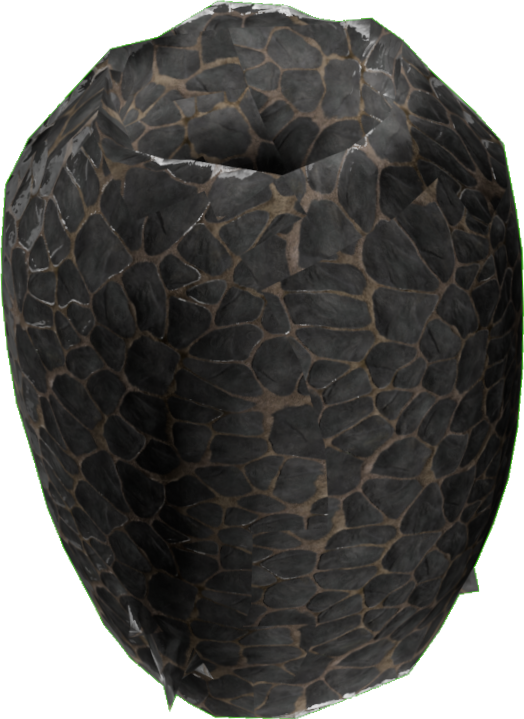} 
            & \imgcell{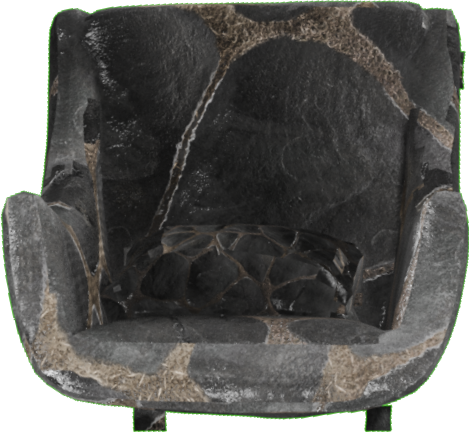} 
         \\
    Leather  
            & \imgcell{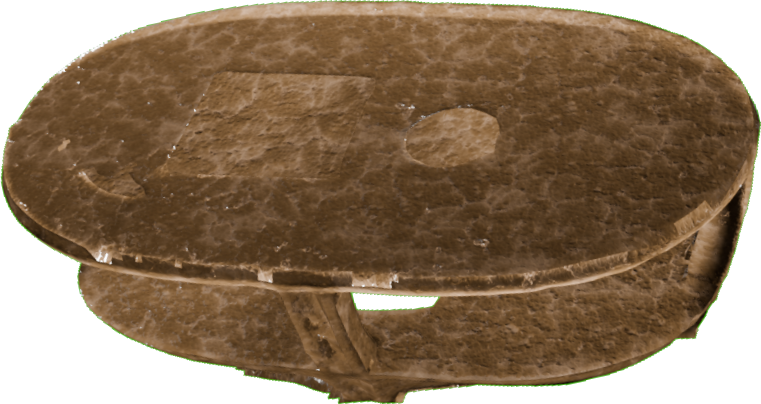}
            & \imgcell{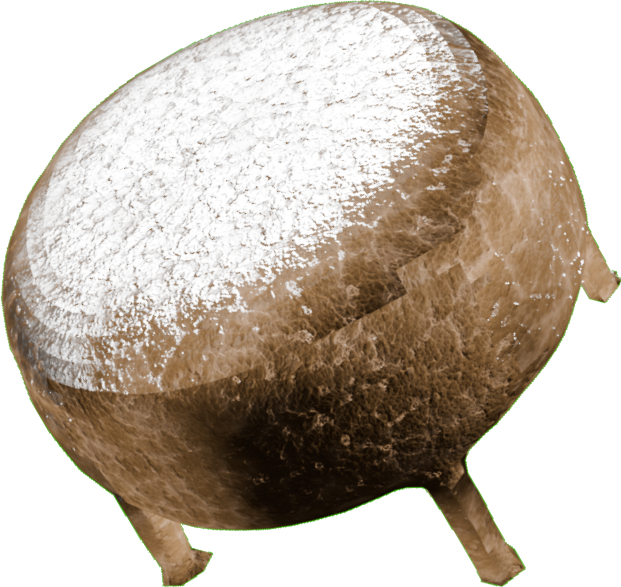}  
            & \imgcell{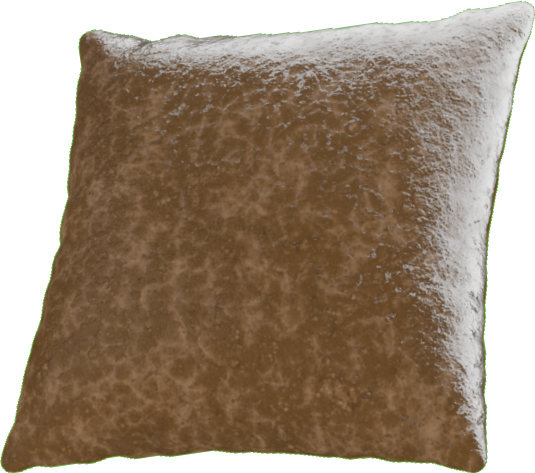} 
            & \imgcell{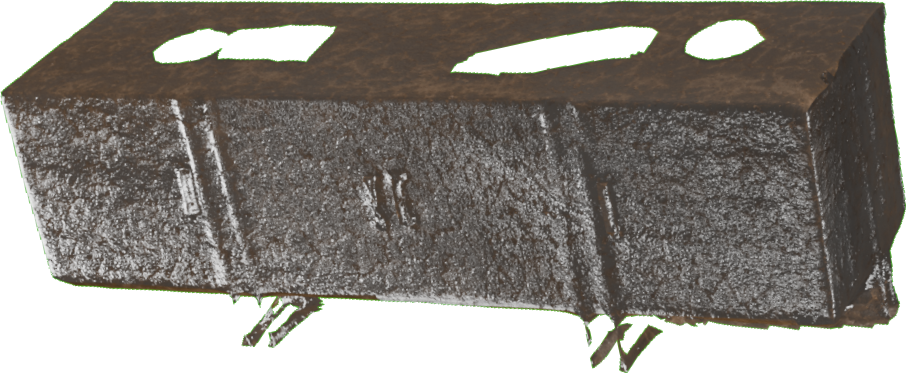} 
            & \imgcell{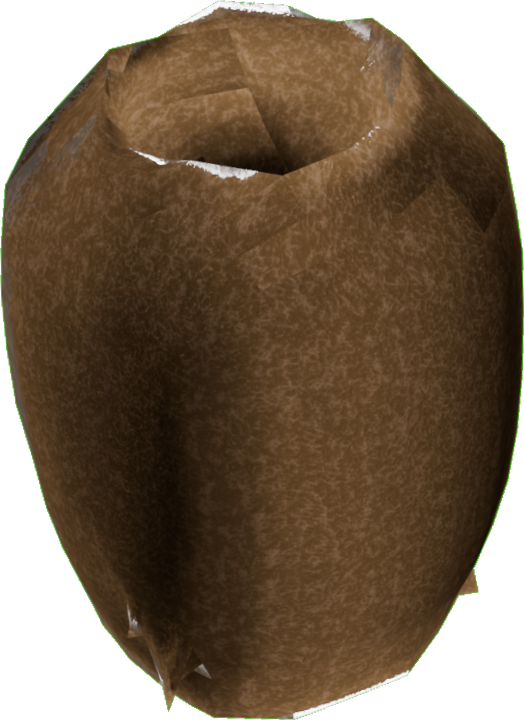} 
            & \imgcell{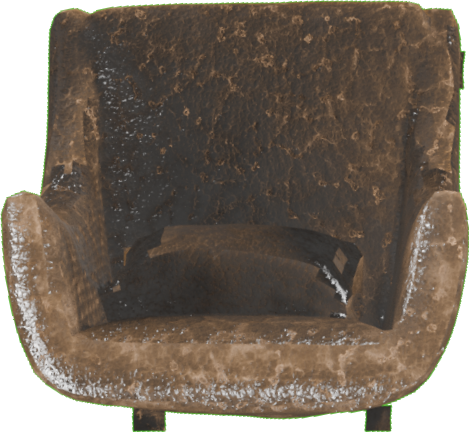} 
         \\
    Fabric  
            & \imgcell{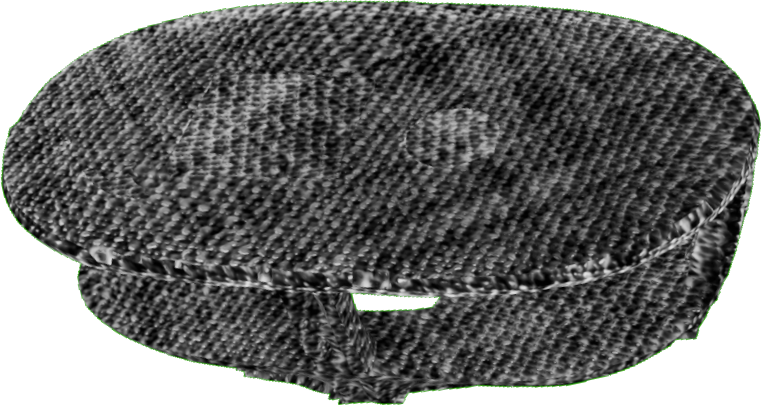}
            & \imgcell{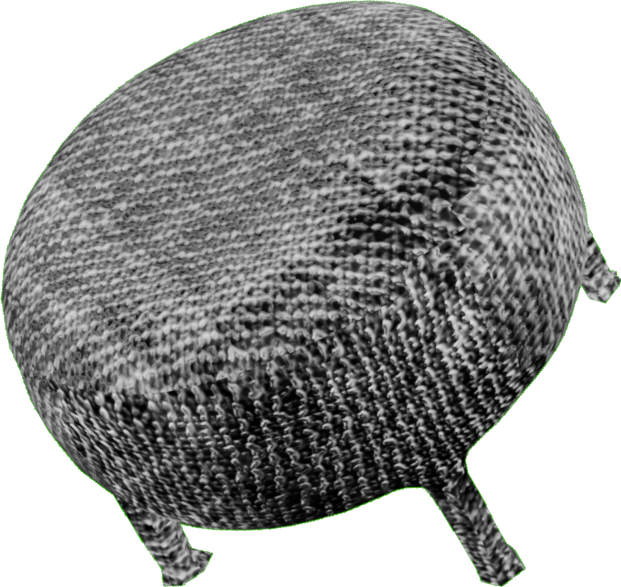}  
            & \imgcell{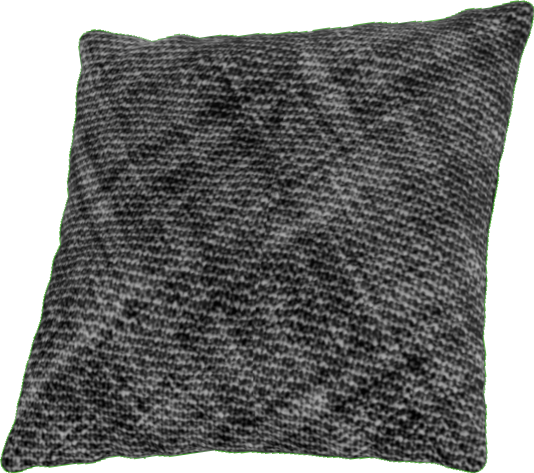} 
            & \imgcell{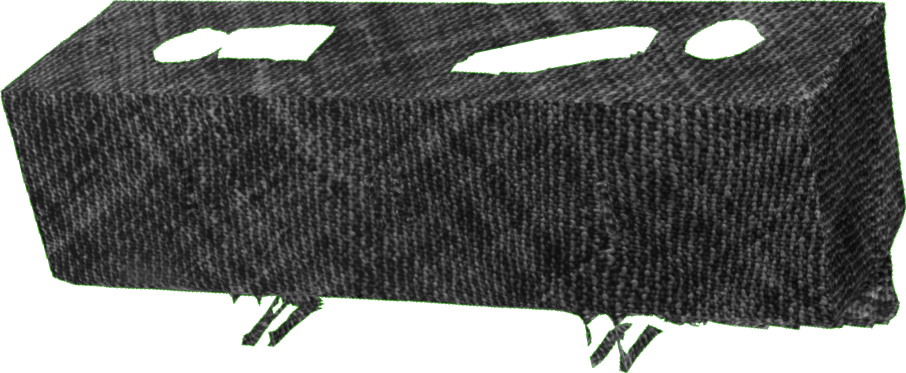} 
            & \imgcell{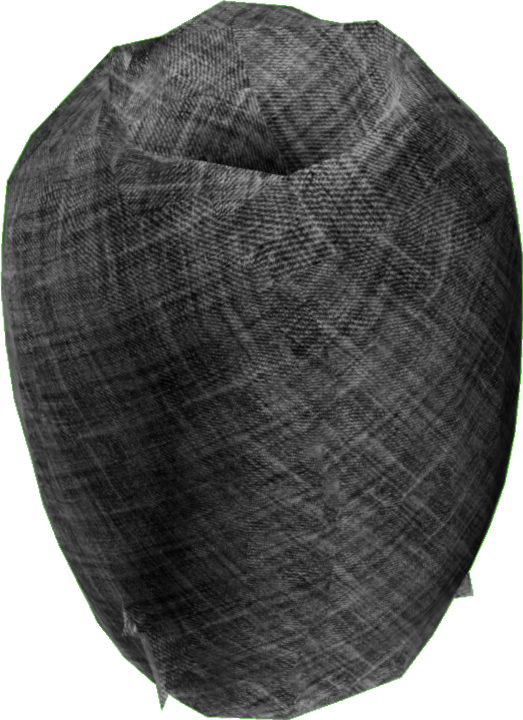} 
            & \imgcell{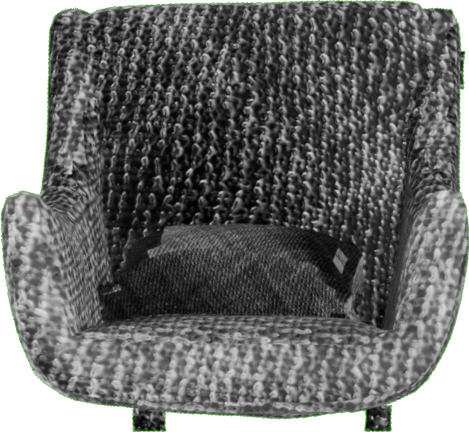} 

       \\
    \bottomrule
  \end{tabularx}
  \label{tab:input_table}
  \vspace{-20pt}
\end{table}

\begin{sidewaystable}
  \centering
  \caption{Benchmark for basecolor  ResNet-50 \& Swin-T}
  \setlength{\tabcolsep}{1pt}
  \renewcommand{\arraystretch}{1.2}

  \begin{tabularx}{1.0\textwidth}{>{\raggedright}p{2cm} p{2.5cm} p{2.5cm} p{2cm}p{2cm}p{2cm} p{2.5cm} p{2cm}p{2cm}p{2cm}}
    \toprule
       \textbf{Material} & \textbf{GroundTruth} & \textbf{ResNet-50} & \textbf{RMSE} & \textbf{SSIM} & \textbf{SAM} & \textbf{Swin-T} & \textbf{RMSE} & \textbf{SSIM} & \textbf{SAM}\\ \midrule
    Wood    
        & \imgcell{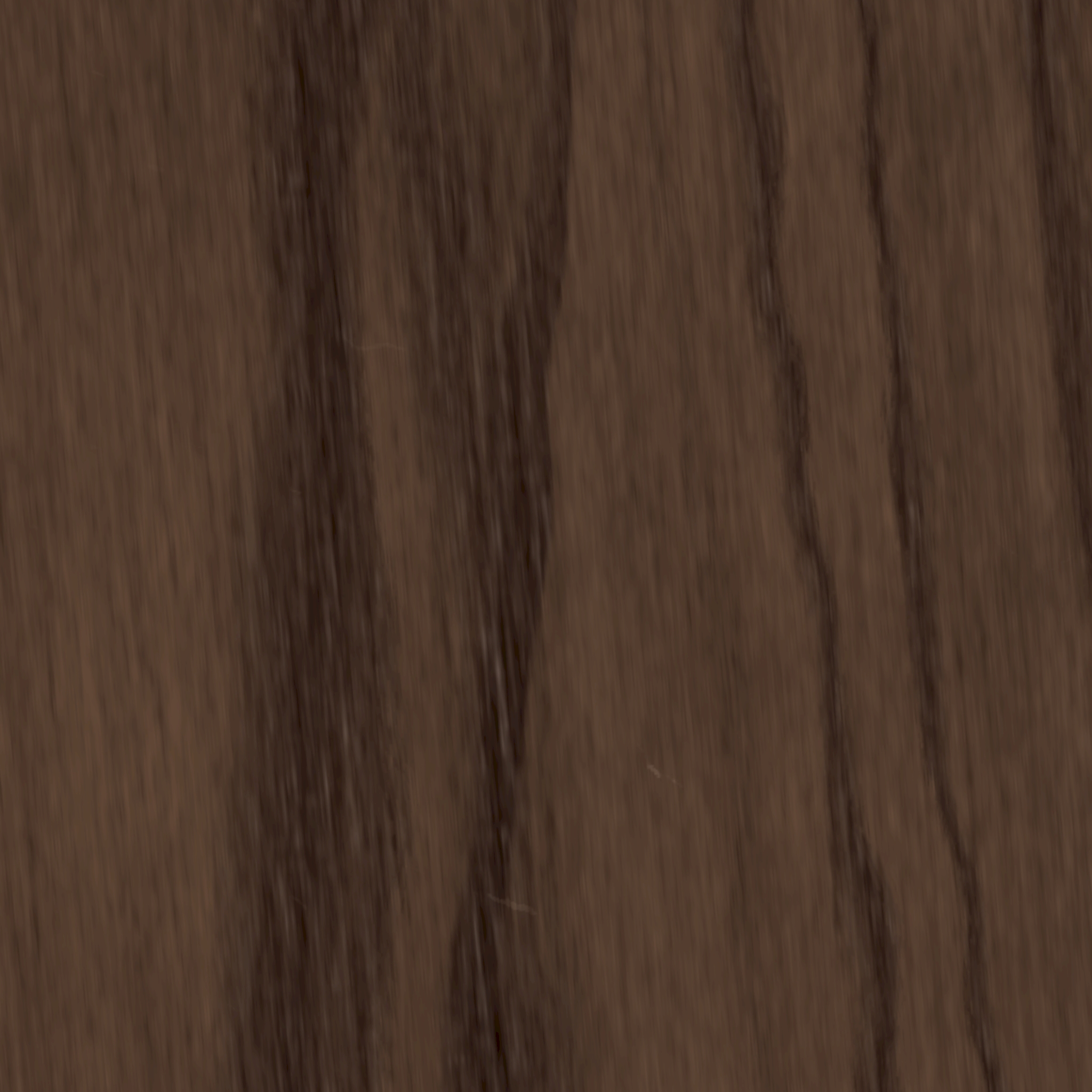}
        & \imgcell{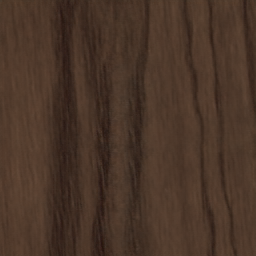}  
        & 0.0025 & 0.9883 & 87.70
        & \imgcell{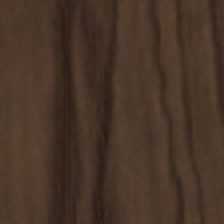}
        & 0.0009 & 0.9990 & 87.63
        \\
    Concrete 
        & \imgcell{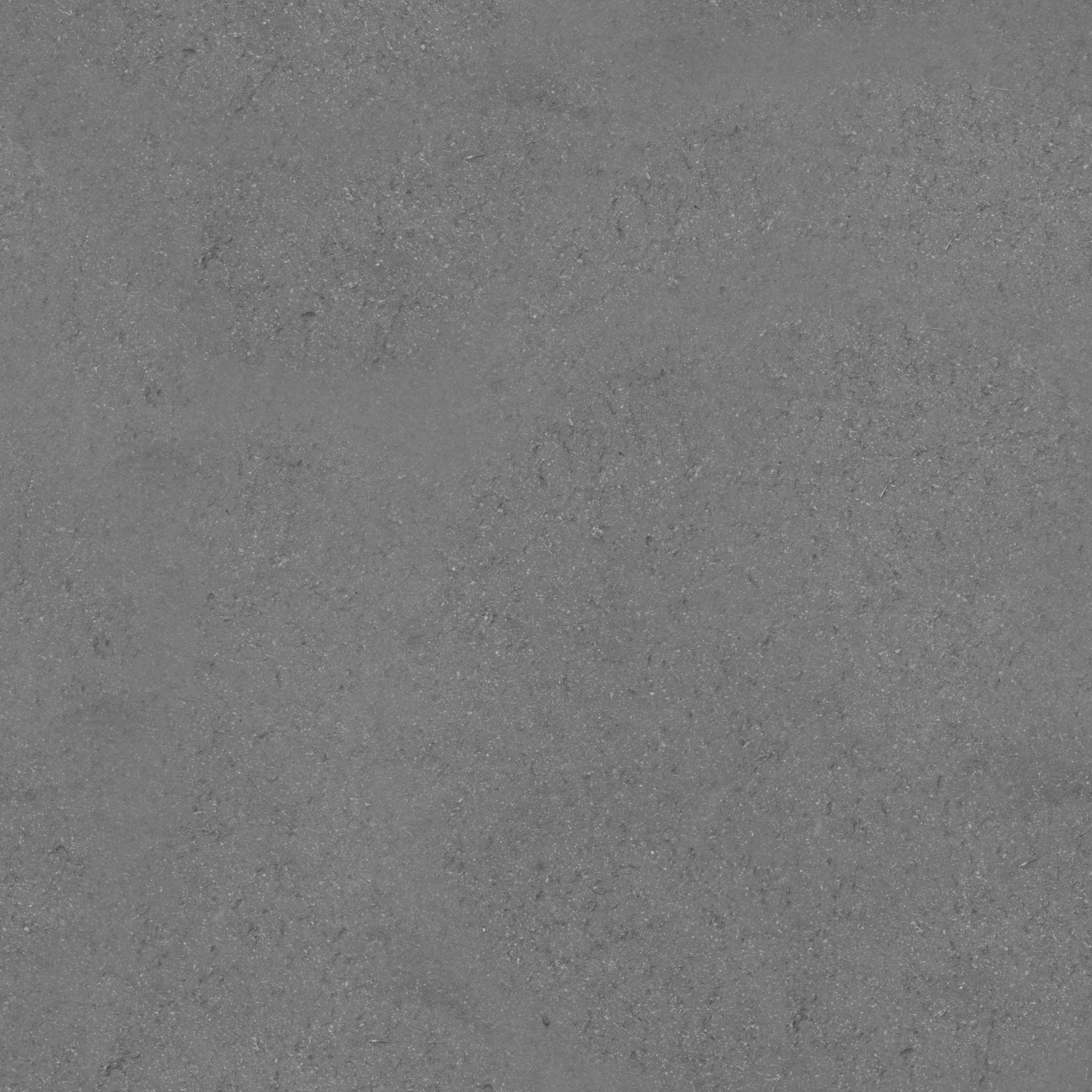}
        & \imgcell{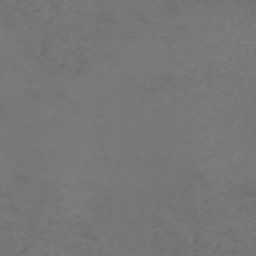} 
        & 0.0009 & 0.9992 & 89.53 
        & \imgcell{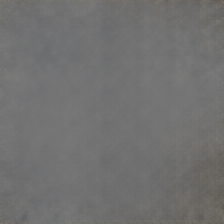}
        & 0.0023 & 0.9967 & 89.45 
        \\
    Plastic 
        & \imgcell{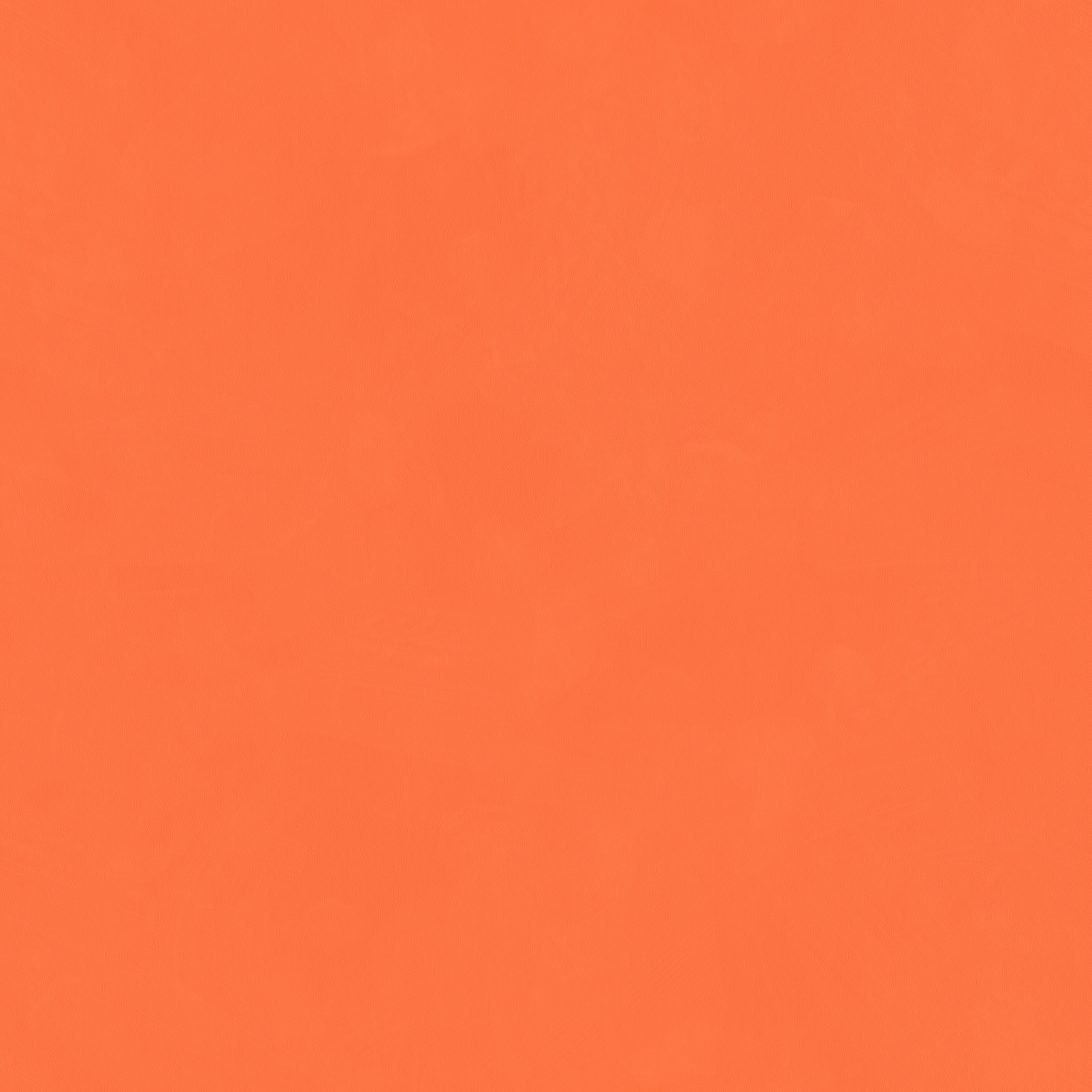}
        & \imgcell{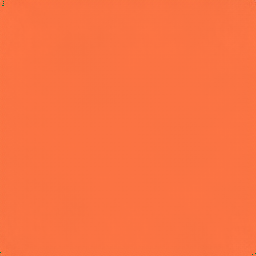}  
        & 0.0005 & 0.9998 & 89.84 
        & \imgcell{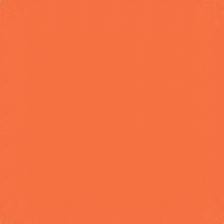}
        & 0.0010 & 0.9997 & 89.82
        \\
    Stone  
        & \imgcell{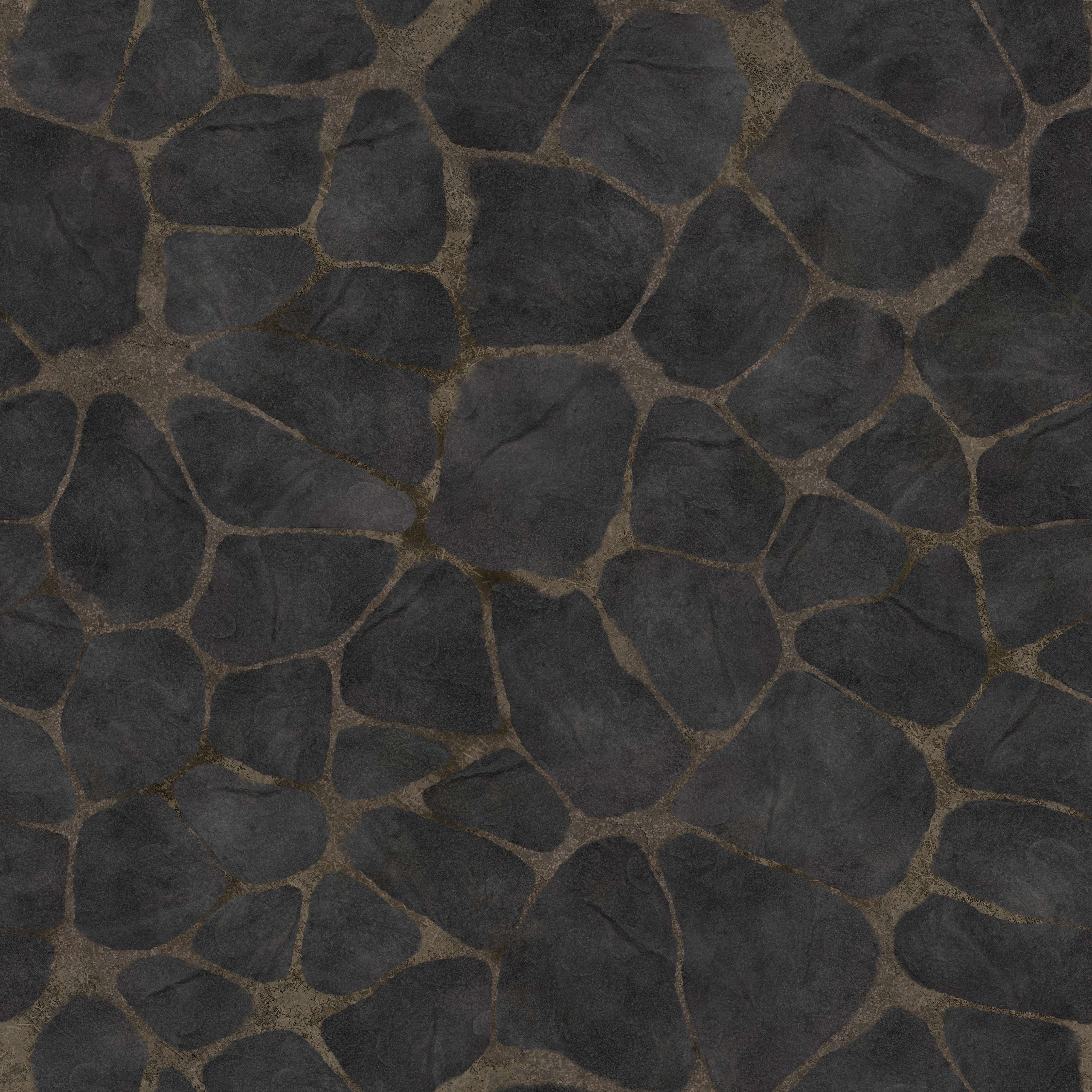}
        & \imgcell{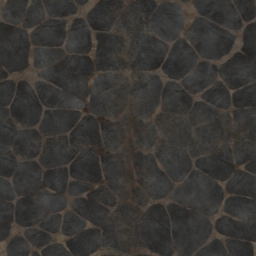}
        & 0.0028 & 0.9878 & 87.42 
        & \imgcell{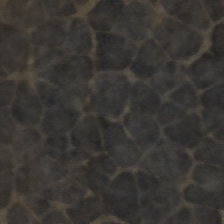}
        & 0.0013 & 0.9975 & 87.38 
        \\
    Leather  
        & \imgcell{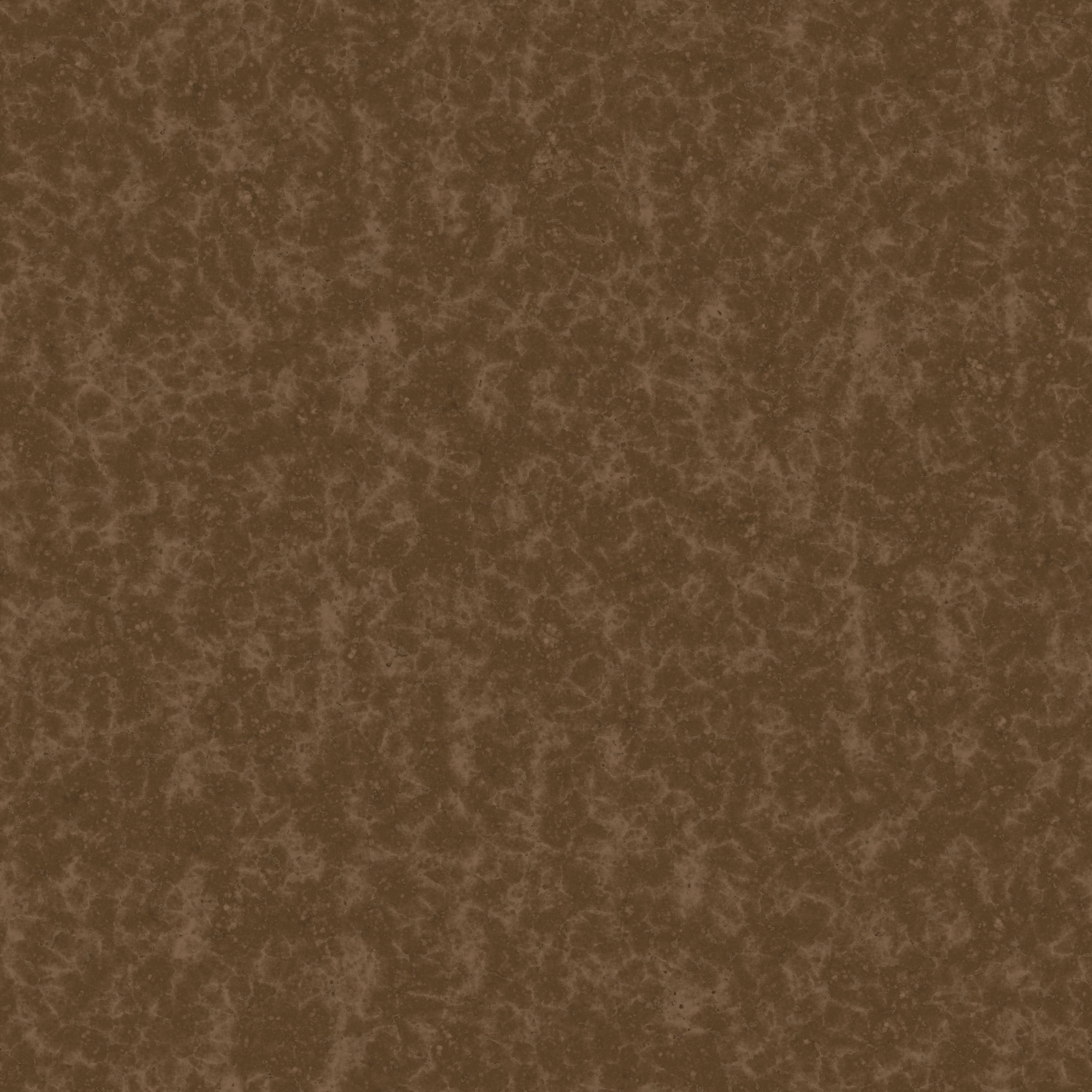}
        & \imgcell{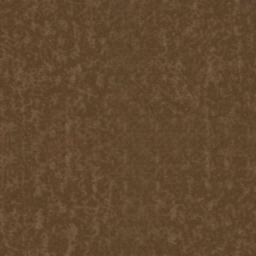}    
        & 0.0020 & 0.9950 & 88.71 
        & \imgcell{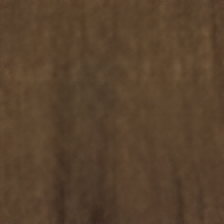}
        & 0.0023 & 0.9931 & 88.59
        \\
    Fabric  
        & \imgcell{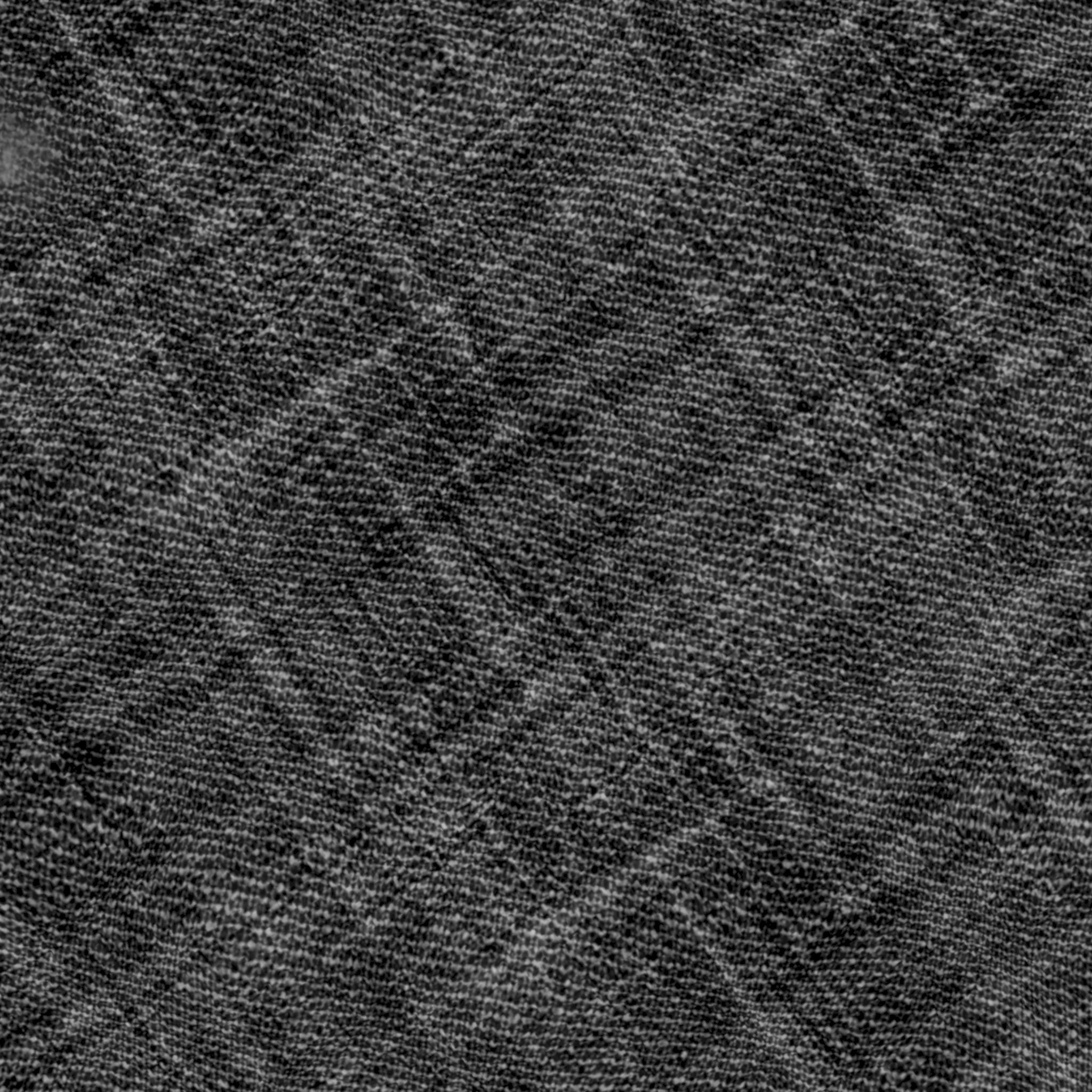}
        & \imgcell{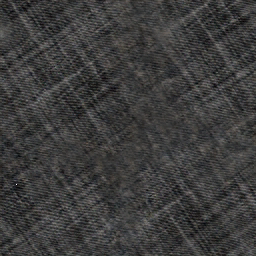} 
        & 0.0066 & 0.9473 & 87.86
        & \imgcell{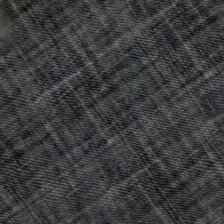}
        & 0.0023 & 0.9938 & 87.12
        \\
    \bottomrule
  \end{tabularx}
  \label{tab:mixed_table}
\end{sidewaystable}

\section{Conclusion and future work}
\label{sec:conclusions}

We have introduced \textsc{MatPredict}, the first dataset that
\emph{factorises material diversity from object geometry}: for every
foreground mesh in Replica we generate multiple photorealistic copies
whose material stack is drawn from the 4 000\,+ entries of
MatSynth.  A physically–motivated pipeline—density–controlled UV
unwrap, UDIM packing, stratified camera shell, and size–aware lighting
rig—yields 18 object categories, 14 material classes and For each object–material pair, we render 512 high-resolution screenshots spanning diverse viewpoints.

On top of the dataset we release a \emph{four-model benchmark}
(UNet-no-skip, ResNet-50, Swin-T, ConvNeXt-T) that learns to regress
\textbf{basecolour \& roughness} ( Table~\ref{tab:mixed_table} and Appendix Table ~\ref{tab:mixed_table_roughness})maps from a single crop.  The shared
encoder–decoder design requires only a $1{\times}1$ head change to
scale to additional channels, and we verify this extensibility on a
two-layer (basecolour + roughness) variant without
changing any other hyper-parameter. Evaluation is reported using three image-similarity metrics spanning the error, perceptual, and spectral families; formal definitions are given in
App.~\ref{app:metrics}, Table~\ref{tab:metric-formulas}.

\textbf{Dataset novelty:}
\textbf{(i) Large-scale synthetic corpus:} we procedurally render a vast set
of object–material combinations in Blender, yielding photo-realistic data
that enables robots to \emph{visually recognise material properties and plan
manipulation} from camera input alone.  
\textbf{(ii) Illumination diversity:} the dataset covers directional, area,
spot and HDR environment lights, so computer-vision models trained on our
images exhibit \emph{enhanced robustness to varying illumination} at
inference time.

\textbf{Limitations:}
Although \textsc{MatPredict} narrows the gap between synthetic and real‐world
captures, several shortcomings remain.  
(i)~The visual fidelity of our renders—particularly the global illumination
and fine caustics produced by transparent media—is still inferior to that of
datasets photographed in real environments; this domain gap may limit final
performance when the trained model is deployed on raw camera frames.  
(ii)~Every mesh in the current release is rendered with a \emph{single,
spatially uniform} material assignment.  Real household objects often exhibit
complex material compositions (e.g.\ a wooden table with a metal frame and
plastic feet), and such heterogeneity may confuse a robot that has only seen
uniform exemplars.  
(iii)~Our benchmark presently targets only two physical layers—
\emph{base-colour} and \emph{roughness}.  Practical manipulation requires
additional properties such as metallicity, normal/displacement maps,
transparency and compliance; predicting those remains future work.

\bibliographystyle{plain}
\bibliography{biblio}

\begin{thebibliography}{10}

\bibitem{RTR4}
Tomas Akenine-M{\"o}ller, Eric Haines, Naty Hoffman, Angelo Pesce, Micha{\l}
  Iwanicki, and S{\'e}bastien Hillaire.
\newblock {\em Real-Time Rendering (4th Edition)}.
\newblock A K Peters / CRC Press, Boca Raton, FL, 2018.

\bibitem{arto2019similarity}
Manuel~Lagunas Arto, Sandra Malpica, Ana Serrano, Elena Garces, Diego
  Gutierrez, and Belen Masia.
\newblock A similarity measure for material appearance.
\newblock {\em Jornada de J{\'o}venes Investigadores del I3A}, 7, 2019.

\bibitem{BlenderSmartUV}
{Blender Documentation Team}.
\newblock {\textit{Mapping Types — Smart UV Project}}.
\newblock {\emph{Blender Manual}}. Blender Foundation, 2024.
\newblock Accessed May 16, 2025.

\bibitem{bruna2013invariant}
Joan Bruna and St{\'e}phane Mallat.
\newblock Invariant scattering convolution networks.
\newblock {\em IEEE transactions on pattern analysis and machine intelligence},
  35(8):1872--1886, 2013.

\bibitem{chan2015pcanet}
Tsung-Han Chan, Kui Jia, Shenghua Gao, Jiwen Lu, Zinan Zeng, and Yi~Ma.
\newblock Pcanet: A simple deep learning baseline for image classification?
\newblock {\em IEEE transactions on image processing}, 24(12):5017--5032, 2015.

\bibitem{Blender2018modelling}
Blender~Online Community.
\newblock {\em Blender - a 3D modelling and rendering package}.
\newblock Blender Foundation, Stichting Blender Foundation, Amsterdam, 2018.

\bibitem{fleming2014visual}
Roland~W Fleming.
\newblock Visual perception of materials and their properties.
\newblock {\em Vision research}, 94:62--75, 2014.

\bibitem{fleming2013perceptual}
Roland~W Fleming, Christiane Wiebel, and Karl Gegenfurtner.
\newblock Perceptual qualities and material classes.
\newblock {\em Journal of vision}, 13(8):9--9, 2013.

\bibitem{He2016ResNet}
Kaiming He, Xiangyu Zhang, Shaoqing Ren, and Jian Sun.
\newblock Deep residual learning for image recognition.
\newblock In {\em IEEE Conference on Computer Vision and Pattern Recognition
  (CVPR)}, pages 770--778, 2016.

\bibitem{jakob2022mitsuba3}
Wenzel Jakob, Sébastien Speierer, Nicolas Roussel, Merlin Nimier-David, Delio
  Vicini, Tizian Zeltner, Baptiste Nicolet, Miguel Crespo, Vincent Leroy, and
  Ziyi Zhang.
\newblock Mitsuba 3 renderer, 2022.
\newblock https://mitsuba-renderer.org.

\bibitem{kirillov2023segment}
Alexander Kirillov, Eric Mintun, Nikhila Ravi, Hanzi Mao, Chloe Rolland, Laura
  Gustafson, Tete Xiao, Spencer Whitehead, Alexander~C Berg, Wan-Yen Lo, et~al.
\newblock Segment anything.
\newblock In {\em Proceedings of the IEEE/CVF international conference on
  computer vision}, pages 4015--4026, 2023.

\bibitem{leung2001representing}
Thomas Leung and Jitendra Malik.
\newblock Representing and recognizing the visual appearance of materials using
  three-dimensional textons.
\newblock {\em International journal of computer vision}, 43(1):29--44, 2001.

\bibitem{li2018cgintrinsics}
Zhengqi Li and Noah Snavely.
\newblock Cgintrinsics: Better intrinsic image decomposition through
  physically-based rendering.
\newblock In {\em Proceedings of the European conference on computer vision
  (ECCV)}, pages 371--387, 2018.

\bibitem{Liu2021Swin}
Ze~Liu, Yutong Lin, Yue Cao, Han Hu, Yixuan Wei, Zheng Zhang, Stephen Lin, and
  Baining Guo.
\newblock Swin transformer: Hierarchical vision transformer using shifted
  windows.
\newblock In {\em IEEE/CVF International Conference on Computer Vision (ICCV)},
  pages 10012--10022, 2021.

\bibitem{Liu2022ConvNeXt}
Zhuang Liu, Hanzi Mao, Chao-Yuan Wu, Christoph Feichtenhofer, Trevor Darrell,
  and Saining Xie.
\newblock A convnet for the 2020s.
\newblock In {\em IEEE/CVF Conference on Computer Vision and Pattern
  Recognition (CVPR)}, pages 11976--11986, 2022.

\bibitem{Ronneberger2015UNet}
Olaf Ronneberger, Philipp Fischer, and Thomas Brox.
\newblock U-net: Convolutional networks for biomedical image segmentation.
\newblock In {\em Medical Image Computing and Computer-Assisted Intervention
  (MICCAI)}, volume 9351 of {\em Lecture Notes in Computer Science}, pages
  234--241. Springer, 2015.

\bibitem{schwartz2019recognizing}
Gabriel Schwartz and Ko~Nishino.
\newblock Recognizing material properties from images.
\newblock {\em IEEE transactions on pattern analysis and machine intelligence},
  42(8):1981--1995, 2019.

\bibitem{sengupta2019neural}
Soumyadip Sengupta, Jinwei Gu, Kihwan Kim, Guilin Liu, David~W Jacobs, and Jan
  Kautz.
\newblock Neural inverse rendering of an indoor scene from a single image.
\newblock In {\em Proceedings of the IEEE/CVF International Conference on
  Computer Vision}, pages 8598--8607, 2019.

\bibitem{song2017semantic}
Shuran Song, Fisher Yu, Andy Zeng, Angel~X Chang, Manolis Savva, and Thomas
  Funkhouser.
\newblock Semantic scene completion from a single depth image.
\newblock In {\em Proceedings of the IEEE conference on computer vision and
  pattern recognition}, pages 1746--1754, 2017.

\bibitem{straub2019replica}
Julian Straub, Thomas Whelan, Lingni Ma, Yufan Chen, Erik Wijmans, Simon Green,
  Jakob~J Engel, Raul Mur-Artal, Carl Ren, Shobhit Verma, et~al.
\newblock The replica dataset: A digital replica of indoor spaces.
\newblock {\em arXiv preprint arXiv:1906.05797}, 2019.

\bibitem{tremeau2020deep}
Alain Tr{\'e}meau, Sixiang Xu, and Damien Muselet.
\newblock Deep learning for material recognition: most recent advances and open
  challenges.
\newblock {\em arXiv preprint arXiv:2012.07495}, 2020.

\bibitem{tuceryan1993texture}
Mihran Tuceryan and Anil~K Jain.
\newblock Texture analysis.
\newblock {\em Handbook of pattern recognition and computer vision}, pages
  235--276, 1993.

\bibitem{vecchio2024matsynth}
Giuseppe Vecchio and Valentin Deschaintre.
\newblock Matsynth: A modern pbr materials dataset.
\newblock {\em arXiv preprint arXiv:2401.06056}, 2024.

\bibitem{Williams1983Mipmaps}
Lance Williams.
\newblock Pyramidal parametrics.
\newblock In {\em Proceedings of the 10th Annual Conference on Computer
  Graphics and Interactive Techniques (SIGGRAPH ’83)}, pages 1--11. ACM,
  1983.

\bibitem{zhang2017physically}
Yinda Zhang, Shuran Song, Ersin Yumer, Manolis Savva, Joon-Young Lee, Hailin
  Jin, and Thomas Funkhouser.
\newblock Physically-based rendering for indoor scene understanding using
  convolutional neural networks.
\newblock In {\em Proceedings of the IEEE conference on computer vision and
  pattern recognition}, pages 5287--5295, 2017.

\end{thebibliography}

\newpage
\begin{appendices}
\section{Closed-form definitions of the eight similarity metrics}
\label{app:metrics}

\begin{table*}[!h]
  \centering
  \caption{Analytic expressions and valid ranges of the eight image–image
           metrics reviewed in §\ref{sec:metrics}.
           Here $X,Y\!\in\!\mathbb{R}^{N}$ are the flattened images
           (or colour vectors), $\mu$ and $\sigma$ denote local means and
           standard deviations, $L$ is the dynamic range, and
           $\langle\cdot,\cdot\rangle$ the Euclidean inner product.}
  \label{tab:metric-formulas}
  \setlength{\tabcolsep}{7pt}
  \renewcommand{\arraystretch}{1.25}
  \begin{tabular}{lll}
    \toprule
    Metric & Formula & Range \\
    \midrule
    RMSE  &
      $\displaystyle\sqrt{\frac{1}{N}\sum_{i=1}^{N}(X_i-Y_i)^{2}}$ &
      $[0,\infty)$, lower is better \\
    PSNR  &
      $\displaystyle 10\log_{10}\!\Bigl(\frac{L^{2}}
        {\frac{1}{N}\sum_{i}(X_i-Y_i)^{2}}\Bigr)$               &
      $[0,\infty)$ dB, higher is better \\
    SRE   &
      $\displaystyle 10\log_{10}\!\Bigl(
        \frac{\sum_i X_i^{2}}{\sum_i (X_i-Y_i)^{2}}\Bigr)$       &
      $[0,\infty)$ dB, higher is better \\
    \midrule
    SSIM  &
      $\displaystyle\frac{(2\mu_X\mu_Y+C_{1})(2\sigma_{XY}+C_{2})}
              {(\mu_X^{2}+\mu_Y^{2}+C_{1})
               (\sigma_X^{2}+\sigma_Y^{2}+C_{2})}$               &
      $[-1,1]$, higher is better \\
    FSIM  &
      $\displaystyle\frac{\sum_{p}PC_{p}\,S_{L}(p)\,S_{\varphi}(p)}
              {\sum_{p}PC_{p}}$                                  &
      $[0,1]$, higher is better \\
    UIQ   &
      $\displaystyle\frac{4\mu_X\mu_Y\sigma_{XY}}
              {(\mu_X^{2}+\mu_Y^{2})(\sigma_X^{2}+\sigma_Y^{2})}$ &
      $[-1,1]$, higher is better \\
    \midrule
    SAM   &
      $\displaystyle\arccos\!\Bigl(
        \frac{\langle X,Y\rangle}{\|X\|_{2}\,\|Y\|_{2}}\Bigr)$   &
      $\bigl[0,\tfrac{\pi}{2}\bigr]$ rad, lower is better \\
    ISSM  &
      $\displaystyle\frac{1}{1+\lVert X-Y\rVert_{F}/\lVert X\rVert_{F}}$ &
      $(0,1]$, higher is better \\
    \bottomrule
  \end{tabular}
\end{table*}

\section{Benchmark for roughness}

\begin{sidewaystable}
  \centering
  \caption{Benchmark for roughness ResNet-50 \& Swin-T}
  \setlength{\tabcolsep}{1pt}
  \renewcommand{\arraystretch}{1.2}

  \begin{tabularx}{1.0\textwidth}{>{\raggedright}p{2cm} p{2.5cm} p{2.5cm} p{2cm}p{2cm}p{2cm} p{2.5cm} p{2cm}p{2cm}p{2cm}}
    \toprule
       \textbf{Material} & \textbf{GroundTruth} & \textbf{ResNet-50} & \textbf{RMSE} & \textbf{SSIM} & \textbf{SAM} & \textbf{Swin-T} & \textbf{RMSE} & \textbf{SSIM} & \textbf{SAM}\\ \midrule
    Wood    
        & \imgcell{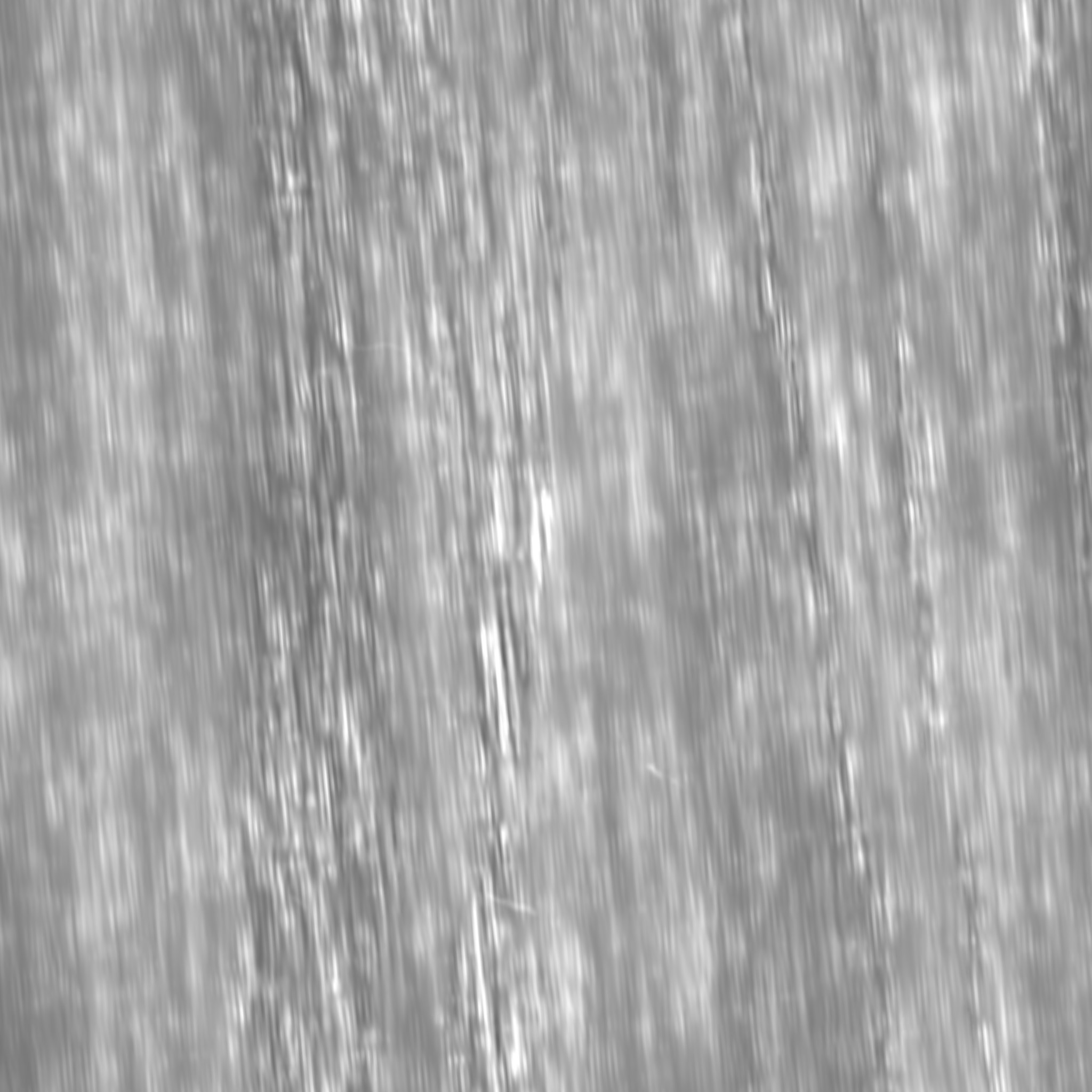}
        & \imgcell{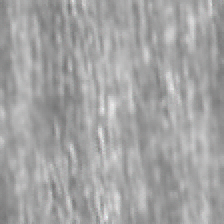}  
        & 0.0020 & 0.9958 & 89.72
        & \imgcell{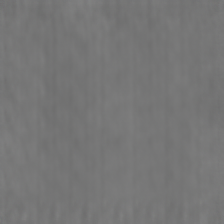} 
        & 0.0124 & 0.9365 & 89.62
        \\
    Concrete 
        & \imgcell{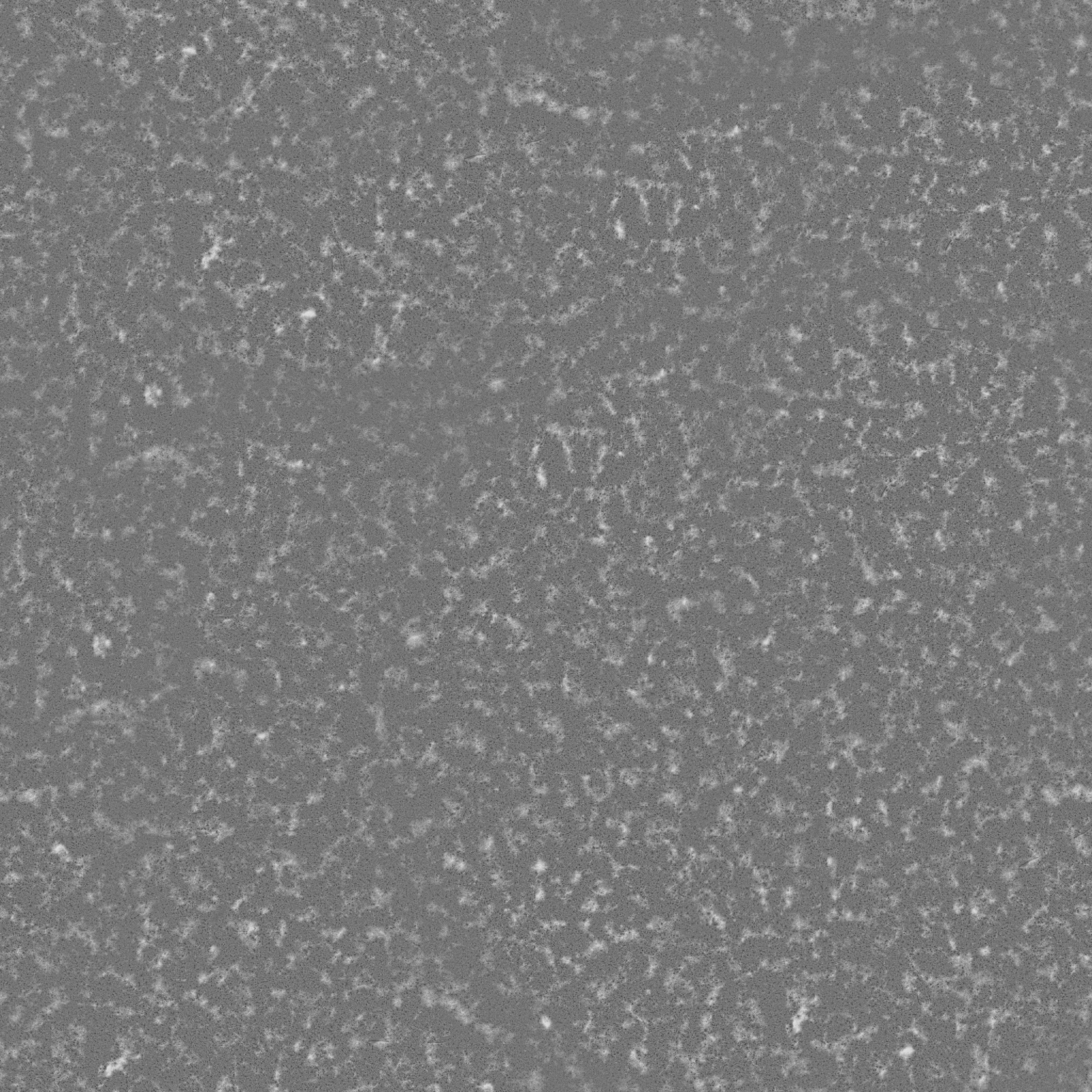}
        & \imgcell{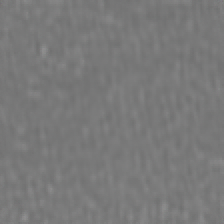} 
        & 0.0.0021 & 0.9952 & 89.52 
        & \imgcell{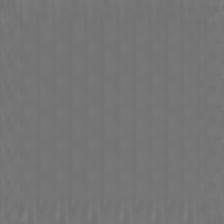} 
        & 0.0026 & 0.9937 & 89.51 
        \\
    Plastic 
        & \imgcell{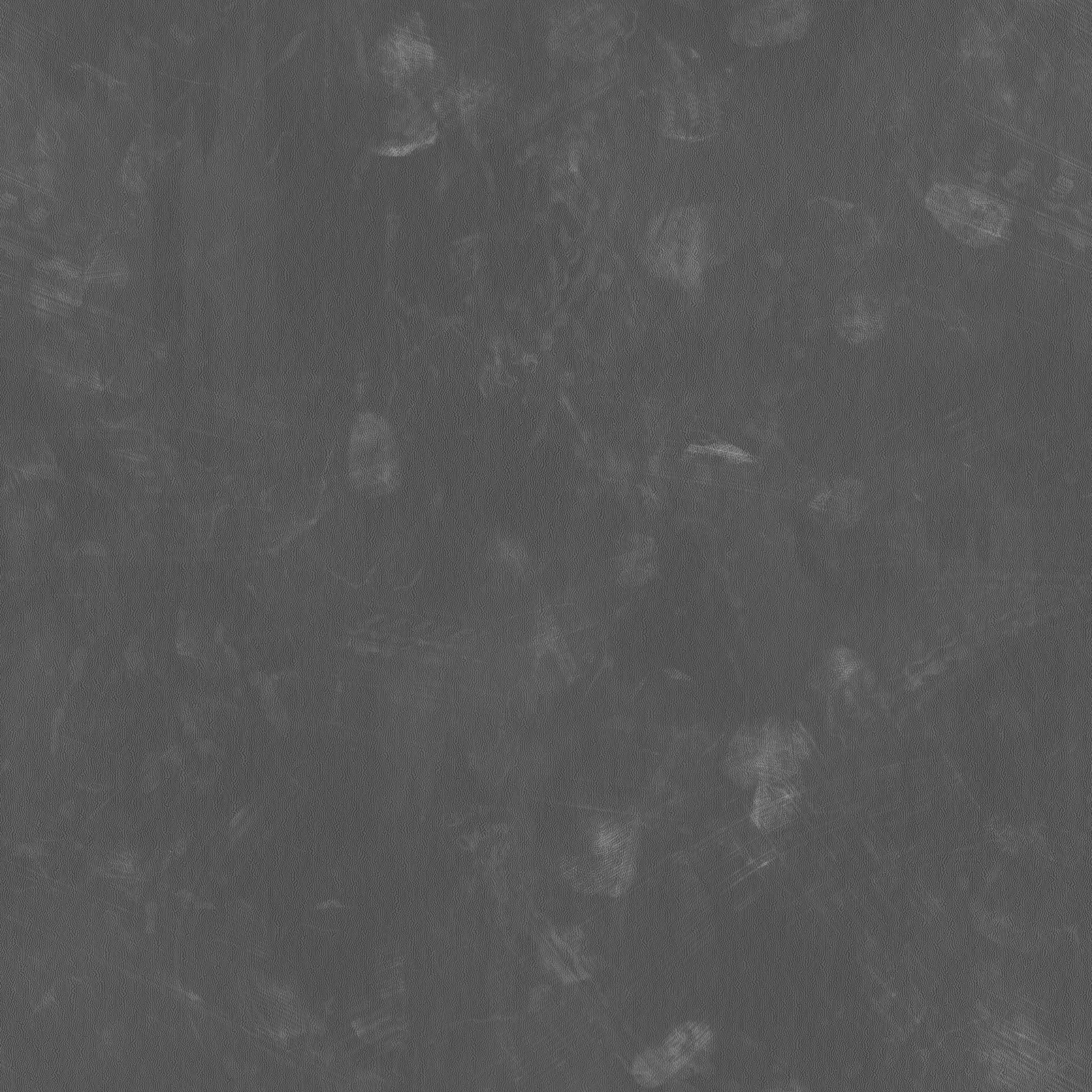}
        & \imgcell{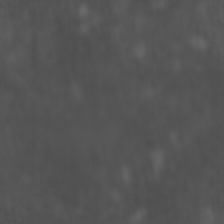}  
        & 0.0007 & 0.9995 & 88.90
        & \imgcell{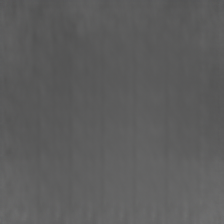} 
        & 0.0019 & 0.9965 & 89.02
        \\
    Stone  
        & \imgcell{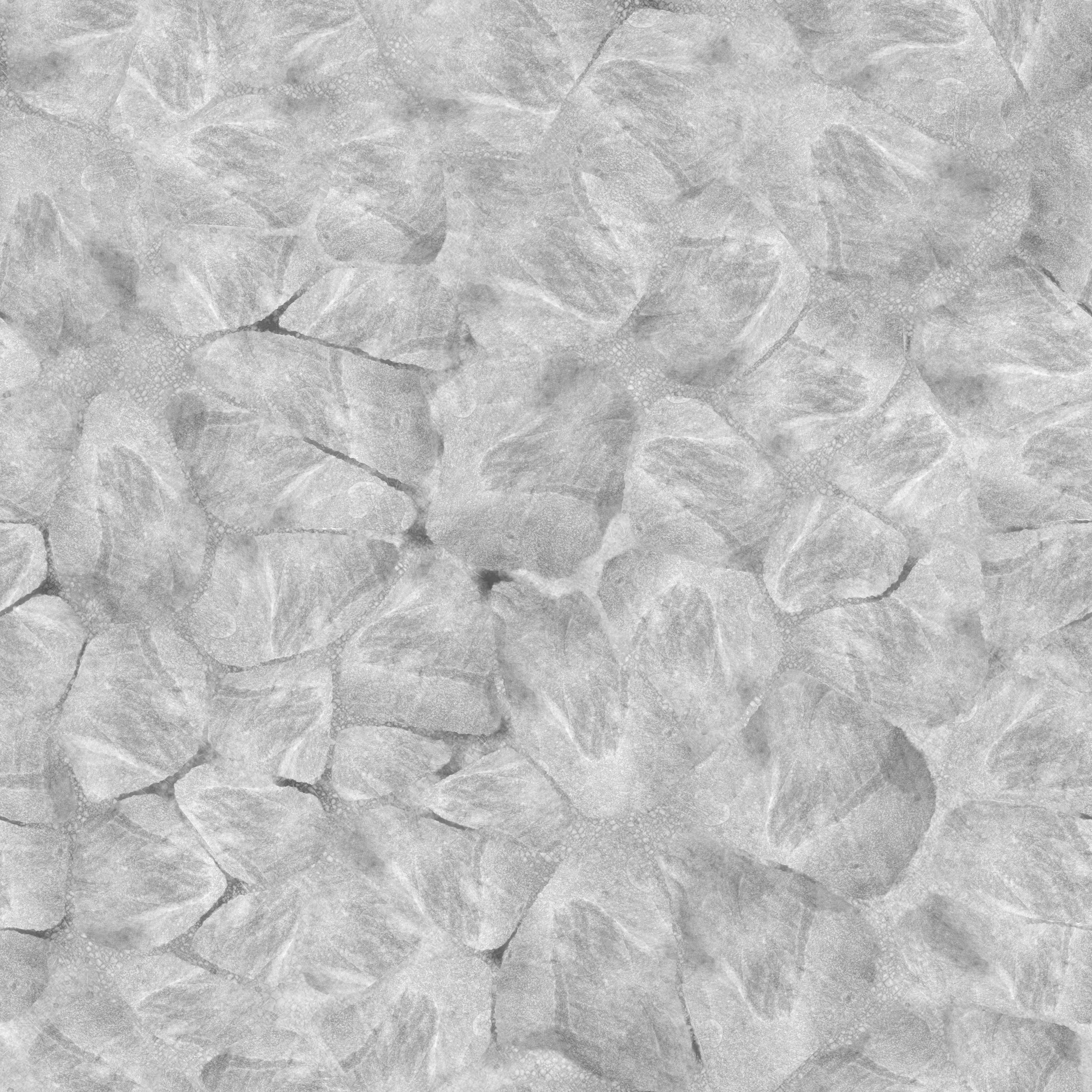}
        & \imgcell{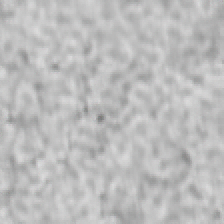}
        & 0.0019 & 0.9962 & 89.79
        & \imgcell{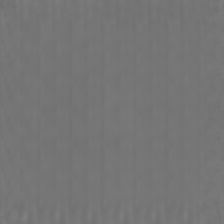} 
        & 0.0182 & 0.8901 & 89.67
        \\
    Leather  
        & \imgcell{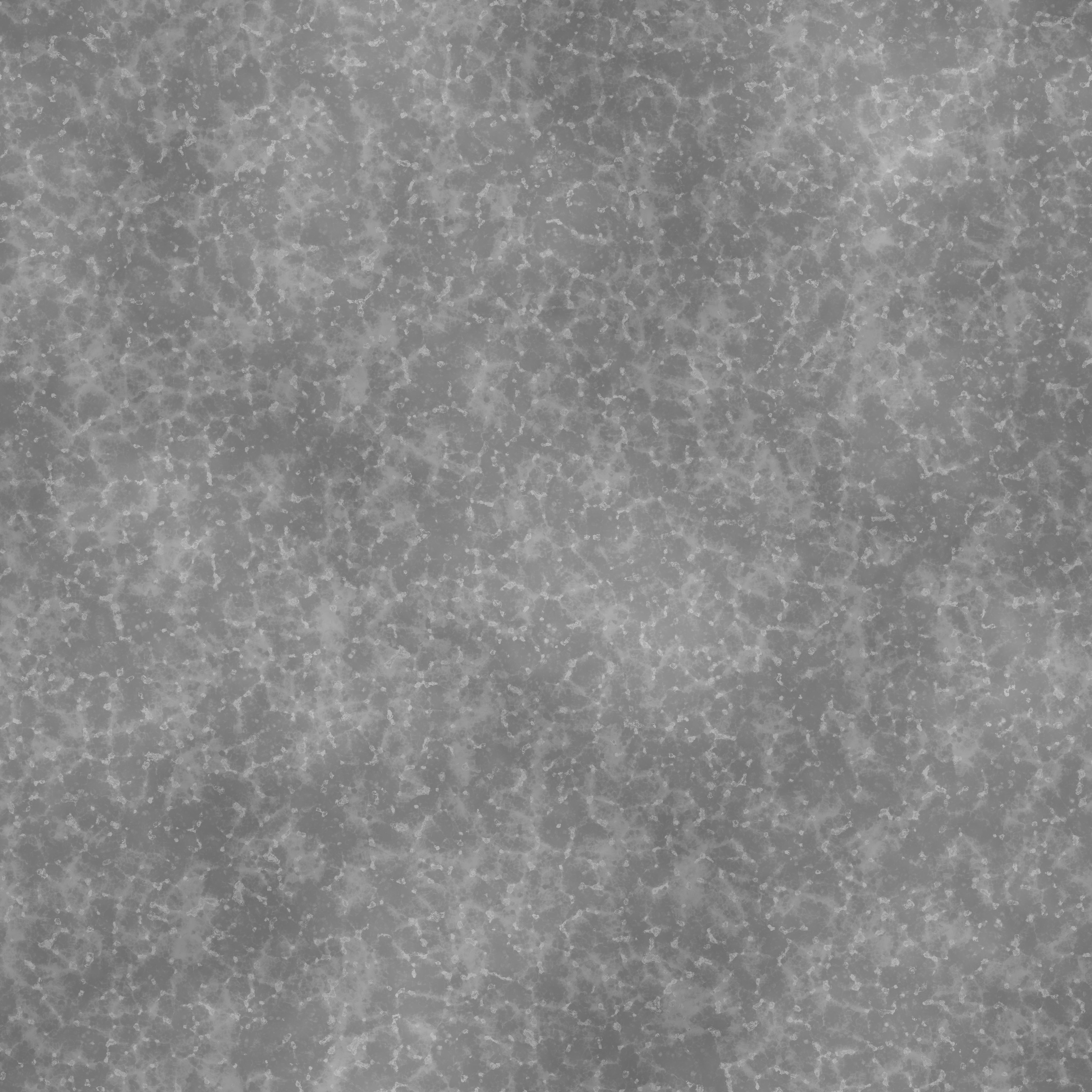}
        & \imgcell{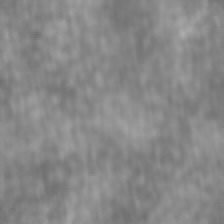}    
        & 0.0016 & 0.9972 & 89.63 
        & \imgcell{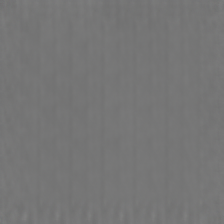} 
        & 0.0054 & 0.9843 & 89.55
        \\
    Fabric  
        & \imgcell{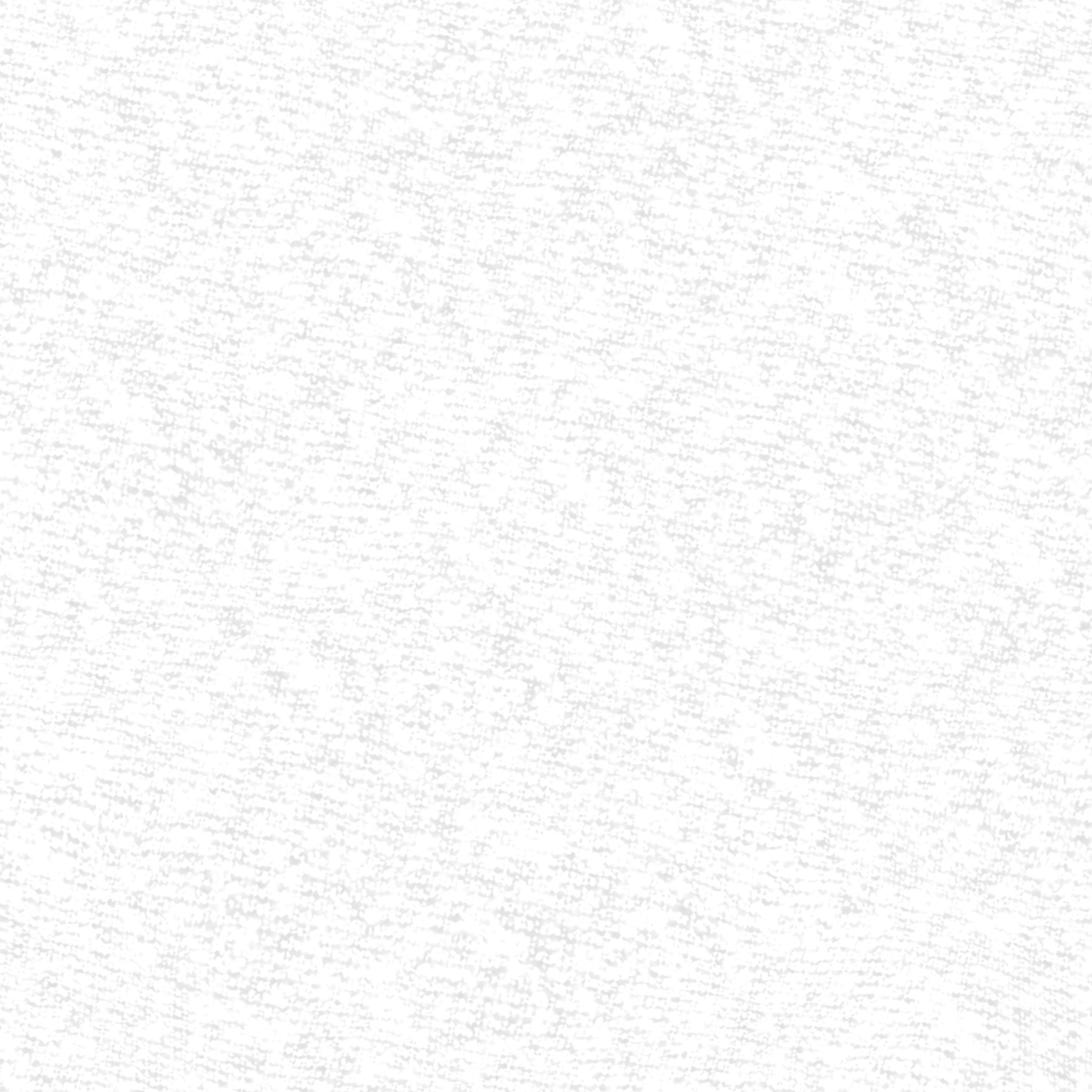}
        & \imgcell{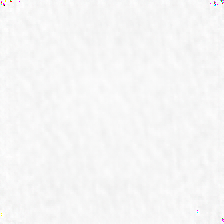} 
        & 0.0017 & 0.9985 & 89.94
        & \imgcell{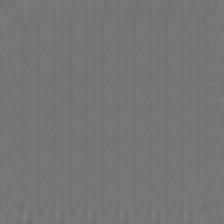} 
        & 0.0322 & 0.7760 & 89.77
        \\
    \bottomrule
  \end{tabularx}
  \label{tab:mixed_table_roughness}
\end{sidewaystable}
\end{appendices}

\end{document}